\documentclass[10pt,twocolumn,letterpaper]{article}

\usepackage{cvpr}              

\usepackage[accsupp]{axessibility}

\usepackage[dvipsnames]{xcolor}

\usepackage{graphicx}
\usepackage{tabularx}
\usepackage{subcaption}

\definecolor{cvprblue}{rgb}{0.21,0.49,0.74}
\usepackage[pagebackref,breaklinks,colorlinks,citecolor=cvprblue]{hyperref}

\def\paperID{*****} 
\def\confName{CVPR}
\def\confYear{2024}

\title{NTIRE 2024 Challenge on Image Super-Resolution ($\times$4): Methods and Results}

\author{
Zheng Chen$^{\dagger}$ \and
Zongwei Wu$^{\dagger}$ \and
Eduard Zamfir$^{\dagger}$ \and
Kai Zhang$^{\dagger}$ \and
Yulun Zhang$^{\dagger,*}$ \and
Radu Timofte$^{\dagger}$ \and
Xiaokang Yang$^{\dagger}$ \and
Hongyuan Yu \and
Cheng Wan \and
Yuxin Hong \and
Zhijuan Huang \and
Yajun Zou \and
Yuan Huang \and
Jiamin Lin \and
Bingnan Han \and
Xianyu Guan \and
Yongsheng Yu \and
Daoan Zhang \and
Xuanwu Yin \and
Kunlong Zuo \and
Jinhua Hao \and
Kai Zhao \and
Kun Yuan \and
Ming Sun \and
Chao Zhou \and
Hongyu An \and
Xinfeng Zhang \and
Zhiyuan Song \and
Ziyue Dong \and
Qing Zhao \and
Xiaogang Xu \and
Pengxu Wei \and
Zhi-chao Dou \and
Gui-ling Wang \and
Chih-Chung Hsu \and
Chia-Ming Lee \and
Yi-Shiuan Chou \and
Cansu Korkmaz \and
A. Murat Tekalp \and
Yubin Wei \and
Xiaole Yan \and
Binren Li \and
Haonan Chen \and
Siqi Zhang \and
Sihan Chen \and
Amogh Joshi \and
Nikhil Akalwadi \and
Sampada Malagi \and
Palani Yashaswini \and
Chaitra Desai \and
Ramesh Ashok Tabib \and
Ujwala Patil \and
Uma Mudenagudi \and
Anjali Sarvaiya \and
Pooja Choksy \and
Jagrit Joshi \and
Shubh Kawa \and
Kishor Upla \and
Sushrut Patwardhan \and
Raghavendra Ramachandra \and
Sadat Hossain \and
Geongi Park \and
S. M. Nadim Uddin \and
Hao Xu \and
Yanhui Guo \and
Aman Urumbekov \and
Xingzhuo Yan \and
Wei Hao \and
Minghan Fu \and
Isaac Orais \and
Samuel Smith \and
Ying Liu \and
Wangwang Jia \and
Qisheng Xu \and
Kele Xu \and
Weijun Yuan \and
Zhan Li \and
Wenqin Kuang \and
Ruijin Guan \and
RutingDeng \and
Zhao Zhang \and
Bo Wang \and
Suiyi Zhao \and
Yan Luo \and
Yanyan Wei \and
Asif Hussain Khan \and
Christian Micheloni \and
Niki Martinel
}

\begin{document}

\maketitle

\let\thefootnote\relax\footnotetext{$^{\dagger}$ Zheng Chen, Zongwei Wu, Eduard Zamfir, Kai Zhang, Yulun Zhang, Radu Timofte, and Xiaokang Yang are the challenge organizers, while the other authors participated in the challenge. $^{*}$ Corresponding author: Yulun Zhang. Appendix~\ref{sec:app_teams} contains the authors' teams and affiliations. NTIRE 2024 webpage: \url{https://cvlai.net/ntire/2024}. Code: \url{https://github.com/zhengchen1999/NTIRE2024_ImageSR_x4}.}

\begin{abstract}
This paper reviews the NTIRE 2024 challenge on image super-resolution ($\times$4), highlighting the solutions proposed and the outcomes obtained. The challenge involves generating corresponding high-resolution (HR) images, magnified by a factor of four, from low-resolution (LR) inputs using prior information. The LR images originate from bicubic downsampling degradation. The aim of the challenge is to obtain designs/solutions with the most advanced SR performance, with no constraints on computational resources (\eg, model size and FLOPs) or training data. The track of this challenge assesses performance with the PSNR metric on the DIV2K testing dataset. The competition attracted 199 registrants, with 20 teams submitting valid entries. This collective endeavour not only pushes the boundaries of performance in single-image SR but also offers a comprehensive overview of current trends in this field.
\vspace{-3.mm}
\end{abstract}

\setlength{\abovedisplayskip}{1pt}
\setlength{\belowdisplayskip}{1pt}

\section{Introduction}
\vspace{1.5mm}
Single image super-resolution (SR) involves the recovery of high-resolution (HR) images from single low-resolution (LR) inputs subjected to a certain degradation process. Image SR is applicable across a variety of visual tasks and application scenarios~\cite{liu2022blind,wang2020deep}. As such, interest in this task has escalated in both academic and industrial sectors. Recently, a continuous stream of methods has been proposed, particularly emphasizing those based on deep neural networks~\cite{dong2016image,sajjadi2017enhancenet,zhou2020cross,liang2021swinir,chen2022cross,chen2022activating}.

In the classical image SR task, LR images are typically produced through bicubic downsampling degradation. The degradation process results in the loss of considerable high-frequency details. The goal of image SR techniques is to restore these high-frequency components to the greatest extent possible, utilizing prior knowledge~\cite{zhang2018image}. This classical degradation model, compared to more complex degradations such as additional noise or blurring, is more conducive to direct comparison of different image SR methods, allowing for the validation of proposed methods performance. Typically, methods that perform well under this type of degradation can also achieve good performance under other degradations through transfer learning~\cite{khan2022ntire}.

Traditional image SR techniques typically rely on interpolation algorithms or reconstruction-based methods, using local pixel information or mathematical modeling for restoration~\cite{capel2003computer,protter2008generalizing,freeman2002example}. With the advancement of hardware technology, deep neural networks have become increasingly prevalent in SR due to their impressive performance~\cite{dong2016image,zhang2017beyond,lai2017deep,zhang2019rnan,liu2020residual}. Starting with the pioneering work of SRCNN~\cite{dong2016image}, which introduced the convolutional neural network (CNN) to SR, there has been a continuous effort to deepen network layers through the use of residual connections~\cite{he2016deep}, thereby enhancing SR performance. Additionally, techniques such as attention mechanisms~\cite{zhang2018image,niu2020singleHAN} and neural architecture search~\cite{li2021heterogeneity,zoph2018learning} are employed to further improve model performance. Apart from CNNs, the Transformer model, initially proposed for natural language processing (NLP), demonstrates exceptional capabilities in SR~\cite{dosovitskiy2020image,liu2021swin,liang2021swinir,zhang2022efficient,chen2022activating,chen2024recursive}. Through its self-attention mechanism, the Transformer can model relationships between global pixels. Moreover, the recently proposed autoregressive model, Mamba~\cite{gu2023mamba}, utilizes causal sequence modeling to capture long-range dependencies in sequences, further advancing the progress in the SR field.

Beyond meticulously designed network structures, a significant amount of effort has been dedicated to boosting model performance from the training and inference perspective. This encompasses the utilization of large pre-trained models~\cite{chen2023activating}, the acquisition or creation of high-quality image data~\cite{wang2021real,zhang2021designing,chen2022real}, the formulation of varied loss functions~\cite{wei2020component}, and the implementation of intricate training strategies~\cite{zamir2021restormer}. Moreover, ensemble strategies~\cite{timofte2016seven} also demonstrate their efficacy, such as the extensively employed self-ensemble technique~\cite{timofte2016seven}. These initiatives, which are complementary to network design and other aspects, improve image SR performance.

Collaborating with the 2024 New Trends in Image Restoration and Enhancement (NTIRE 2024) workshop, we developed a challenge centred on single image super-resolution ($\times$4). This challenge aims to upscale a single low-resolution image by a factor of four, enhancing the detail level of high-resolution images. The goal is to discover SR solutions that deliver state-of-the-art performance (\eg, PSNR), and to determine prevailing trends in the design of image SR networks.

This challenge is one of the NTIRE 2024 Workshop~\footnote{\url{https://cvlai.net/ntire/2024/}} associated challenges on: dense and non-homogeneous dehazing~\cite{ntire2024dehazing}, night photography rendering~\cite{ntire2024night}, blind compressed image enhancement~\cite{ntire2024compressed}, shadow removal~\cite{ntire2024shadow}, efficient super resolution~\cite{ntire2024efficientsr}, image super resolution ($\times$4) (this challenge), light field image super-resolution~\cite{ntire2024lightfield}, stereo image super-resolution~\cite{ntire2024stereosr}, HR depth from images of specular and transparent surfaces~\cite{ntire2024depth}, bracketing image restoration and enhancement~\cite{ntire2024bracketing}, portrait quality assessment~\cite{ntire2024QA_portrait}, quality assessment for AI-generated content~\cite{ntire2024QA_AI}, restore any image model (RAIM) in the wild~\cite{ntire2024raim}, RAW image super-resolution~\cite{ntire2024rawsr}, short-form UGC video quality assessment~\cite{ntire2024QA_UGC}, low light enhancement~\cite{ntire2024lowlight}, and RAW burst alignment and ISP challenge.

\section{NTIRE 2024 Image Super-Resolution ($\times$4) Challenge}
This challenge is part of the NTIRE 2024 associated challenges, aiming to achieve two main goals: (1) to present a comprehensive review of the recent developments and emerging trends in the image SR domain; and (2) to serve as a forum where academic and industrial practitioners can come together and explore potential collaborations. This section explores the specific details of the challenge.

\subsection{Dataset}

The challenge provides several datasets: DIV2K~\cite{timofte2017ntire}, Flickr2K~\cite{lim2017enhanced}, and LSDIR~\cite{li2023lsdir}. Meanwhile, the usage of extra data is supported. In this challenge, low-resolution versions of images are generated by applying bicubic interpolation with a downsampling factor of $\times$4.

\paragraph{DIV2K.} The DIV2K dataset comprises 1,000 RGB images with 2K resolution. It is organized into three parts: 800 images for training, 100 images for validation, and 100 images for testing. To ensure fairness, the HR images of the DIV2K validation set are hidden from participants during the validation phase. Meanwhile, the test HR images are kept hidden throughout the challenge.

\paragraph{Flickr2K.} The Flickr2K dataset includes 2,650 2K images sourced from the Flickr. These images display a variety of content and quality levels. For this challenge, all images are made available to the contestants.

\paragraph{LSDIR.} The LSDIR dataset features 86,991 high-resolution images sourced from Flickr platform, divided into three segments: 84,991 for training, 1,000 for validation, and 1,000 for testing.

\subsection{Track and Competition}
The objective is to develop a network to generate high-quality results, achieving optimal performance (\ie, PSNR).

\paragraph{Track: Restoration Quality.} Consistent with the approach taken in the previous year~\cite{zhang2023ntire_sr}, teams are evaluated and ranked based on the PSNR values of their enhanced high-resolution images compared to the ground truth high-resolution images from the DIV2K testing dataset.

\paragraph{Challenge Phases.} \textit{(1) Development and Validation Phase:} Contestants gain access to 800 LR/HR training image pairs and 100 LR validation images within the DIV2K dataset. In addition, they are permitted to apply additional data for training. Participants can submit their restored HR images to the Codalab evaluation server to ascertain the PSNR of their model-generated SR images and receive immediate feedback. \textit{(2) Testing Phase:} During this final phase, contestants receive 100 LR test images, while the corresponding HR ground truth images remain concealed. Subsequently, contestants upload their super-resolution results to the Codalab server and submit their code along with a detailed report to the organizers via email. The organizers execute the code for validation and communicate the results to the participants after the challenge concludes.

\paragraph{Evaluation Protocol.} The assessment employs two standard metrics: Peak Signal-to-Noise Ratio (PSNR) and Structural Similarity Index (SSIM), where PSNR is the primary one. During the calculation, a 4-pixel border around each image is discarded, and calculations are performed on the Y channel of the YCbCr color space. For final results, priority is given to the submission on the Codalab server. Results generated from the submitted code are used for reproduction and verification, with minor drops in precision being acceptable. A script for these metric calculations is accessible at \url{https://github.com/zhengchen1999/NTIRE2024_ImageSR_x4}, where the repository further includes the source code and pre-trained models of the entries submitted.

\begin{table}[t]
    \centering
    \footnotesize
    \begin{tabularx}{\columnwidth}{X | c | cc}
    \toprule
        Team & Rank & PSNR (primary) & SSIM  \\
     \midrule
        XiaomiMM & 1 & 31.94 & 0.8778 \\
        SUPSR & 2 & 31.41 & 0.8711\\
        UCAS-SCST & 3 & 31.28 & 0.8666\\
        SYSU-SR & 4 & 31.19 & 0.8660 \\
        Jasmine & 5 & 31.18 & 0.8665\\
        ACVLAB & 6 & 31.18 & 0.8655\\
        mandalinadagi & 7 & 31.13 & 0.8648\\
        SKDADDYS & 8 & 31.11 & 0.8643\\
        KLETech-CEVI & 9 & 31.03 & 0.8633\\
        SVNIT-NTNU & 10 & 31.03 & 0.8633\\
        ResoRevolution & 11 & 31.01 & 0.8647\\
        BetterSR & 12 & 30.97 & 0.8621\\
        Contrast & 13 & 30.69 & 0.8563\\
        BFU-SR & 14 & 30.55 & 0.8560\\
        SCU-VIP-LAB & 15 & 29.78 & 0.8506\\
        Nudter & 16 & 30.17 & 0.8446\\
        JNU-620 & 17 & 30.43 & 0.8426\\
        LVGroup-HFUT & 18 & 29.98 & 0.8380\\
        Uniud & 19 & 29.97 & 0.8440\\
        SVNIT-NTNU-1 & 20 & 29.34 & 0.8199\\
     \bottomrule
    \end{tabularx}
    \caption{\textbf{Results of NTIRE 2024 Image Super-Resolution Challenge.} PSNR and SSIM scores are measured on the DIV2K testing (100) dataset. Team rankings are based primarily on PSNR, with SSIM as the secondary criterion.}
    \label{tab:main_results}
    \vspace{-4.mm}
\end{table}

\section{Challenge Results}
\vspace{1.mm}
Table~\ref{tab:main_results} presents the final rankings and results of the teams. Details on the evaluation methodology can be found in Sec.~\ref{sec:teams}, and the roster of team members is detailed in Appendix~\ref{sec:app_teams}. In this challenge, the XiaomiMM team achieved the highest overall ranking. Additionally, the PSNR values for the top six teams exceeded 31.1 dB. Notably, the results of the top three teams surpass the highest result from the previous year challenge.

\vspace{3.mm}
\subsection{Architectures and main ideas}
\vspace{1.mm}
Throughout the challenge, various innovative techniques were introduced to boost the SR performance. Here, we summarize some of the principal concepts.
\begin{enumerate}
    \item \textbf{Employing pre-trained Transformers is a mainstream approach.} Transformer-based SR methods still achieve complelling reconstruction results, such as SwinIR~\cite{liang2021swinir}, HAT~\cite{chen2023activating} and DAT~\cite{chen2023dual}. Therefore, several teams participating in this challenge have investigated employing large pre-trained models, which are further fine-tuned using different datasets of loss functions. For example, the SYSU-SR team proposed enhancement in the two phases of training and testing.
    
    \item \textbf{Adopting novel Mamba-based design.} Modeling contextual and global information is highly important for improving the reconstruction fidelity of Super-Resolution methods. This year winning team, XiaomiMM, successfully employed the popular Mamba architecture to Super-Resolution, outperforming the second-placed method by $0.53$dB.
    
    \item \textbf{Scaling up high-quality image Super-Resolution datasets.} Recent advancements in other vision domains, driven by the scalability of model size and training data, have sparked the interest of some participants in gathering extensive high-quality image datasets, thus achieving superior performance. The participant SUPSR proposes a high-quality data selection scheme and constructs a dataset of $25$M images.
    
    \item \textbf{Incorporating frequency domain for better detail reconstruction.} Several participants integrate additional components or loss functions in the frequency domain to strengthen the overall recovery of fine details. For example, the competitor UCAS-SCST designs a High Frequency Transformer (HFT) based on the popular HAT~\cite{chen2023activating} model, while the team mandalinadagi employs a Wavelet-based loss function.
    
    \item \textbf{Advanced training strategy also boosts performance.} Certain teams have applied sophisticated training strategies in their methods. For example, participants adjusted the crop size during training or utilized varying crop sizes at different stages of the training process.
    
    \item \textbf{Enhancing performance through image augmentation.} The self-ensemble strategy~\cite{timofte2016seven} has proven to be successful in boosting performance and is extensively utilized. Mirroring the approach of the previous year's winner~\cite{zhang2023ntire_sr}, several teams implemented various augmentation methods during training or testing phases to enhance SR results further.
\end{enumerate}

\vspace{-1.mm}
\subsection{Participants}
\vspace{-1.mm}
This year, the image SR challenge saw 199 registered participants, with 20 teams providing valid submissions. Compared to last year challenge~\cite{zhang2023ntire_sr}, there are more valid submissions (from 15 to 20) and better results. These entries set a new standard for the state-of-the-art in image SR ($\times$4).

\vspace{-1.mm}
\subsection{Fairness}
\vspace{-1.mm}
A set of rules is instituted to maintain the fairness of the competition. \textbf{(1)} The use of DIV2K test HR images for training is prohibited, while the DIV2K test LR images are accessible. \textbf{(2)} Training with additional datasets, \eg, Flickr2K and LSDIR datasets, is permitted. \textbf{(3)} The application of data augmentation techniques during training and testing is deemed a fair practice.

\vspace{-1.mm}
\subsection{Conclusions}
\vspace{-1.mm}
The insights gained from analyzing the results of the image SR challenge are summarized as follows:
\begin{enumerate}
    \item The techniques introduced in this challenge have significantly propelled the progress and practical applications in the image SR sector.
    \item The usage of the Transformer architecture continues to exhibit impressive performance, while Mamba-based approaches exhibit promising potential to explore new directions in architectural design, offering strong reconstruction performance.
    \item Scalability shows significant potential for the Super-Resolution community, underscoring the increasing need for high-quality, extensive datasets. These datasets are particularly vital for the enhancement of large-scale neural networks.
\end{enumerate}

\vspace{-2.mm}
\section*{Acknowledgements}
\vspace{-2.mm}
This work was partially supported by the Humboldt Foundation, Shanghai Municipal Science and Technology Major Project (2021SHZDZX0102), and NSFC (U19B2035). We thank the NTIRE 2024 sponsors: Meta Reality Labs, OPPO, KuaiShou, Huawei, and University of W\"urzburg (Computer Vision Lab).

\vspace{-1.mm}
\section{Challenge Methods and Teams}
\label{sec:teams}
\vspace{-1.mm}

\begin{figure}
    \centering
    \includegraphics[height=1\columnwidth, angle=90]{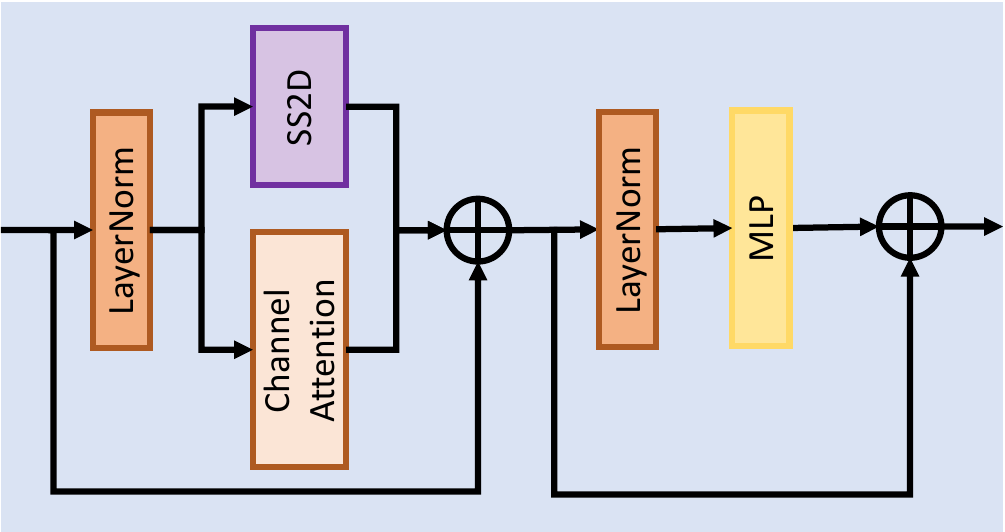}
    \vspace{-4.mm}
    \caption{\textbf{Team XiaomiMM}}
    \label{fig:team1_1}
    \vspace{-4.mm}
\end{figure}

\subsection{XiaomiMM}
\vspace{-1.mm}
\paragraph{Description.} The solution proposed by the XiaomiMM is illustrated in \cref{fig:team1_1}. The characteristic of Mamba~\cite{liu2024vmamba} is its ability to model long-range dependencies of long sequences, which is likely due to its parametric approach that enables Mamba to store information of long sequences. However, Mamba is an autoregressive model, which typically has unidirectionality, such as good temporal properties and causal sequence modeling. Compared to the Transformer~\cite{vaswani2017attention}, it cannot model the relationships between sequence elements. The Transformer has shown strong advantages in various tasks, but it is not good at handling long sequence information. The characteristics of Mamba and Transformer are highly complementary, for which the authors designed the SSFormer (State Space Transformer) block. The Super-Resolution (SR) task~\cite{wan2024swift} is indeed a pixel-intensive task because it aims to recover high-resolution (HR) details from low-resolution (LR) images. In this process, the model needs to perform dense calculations at each pixel point to predict and generate new pixel points in higher resolution images, so modeling the contextual relationship of pixel points in the super-resolution task is more important. Based on this, the authors introduced the SSFormer Block into the super-resolution task and built the MambaSR model. The network structure of MambaSR is based on HAT~\cite{chen2023activating}, and the authors replaced all the Hybrid Attention Blocks (HAB) of HAT with SSFormer Blocks, achieving the best performance.

\begin{figure*}[t]
    \centering
    \includegraphics[width=\textwidth]{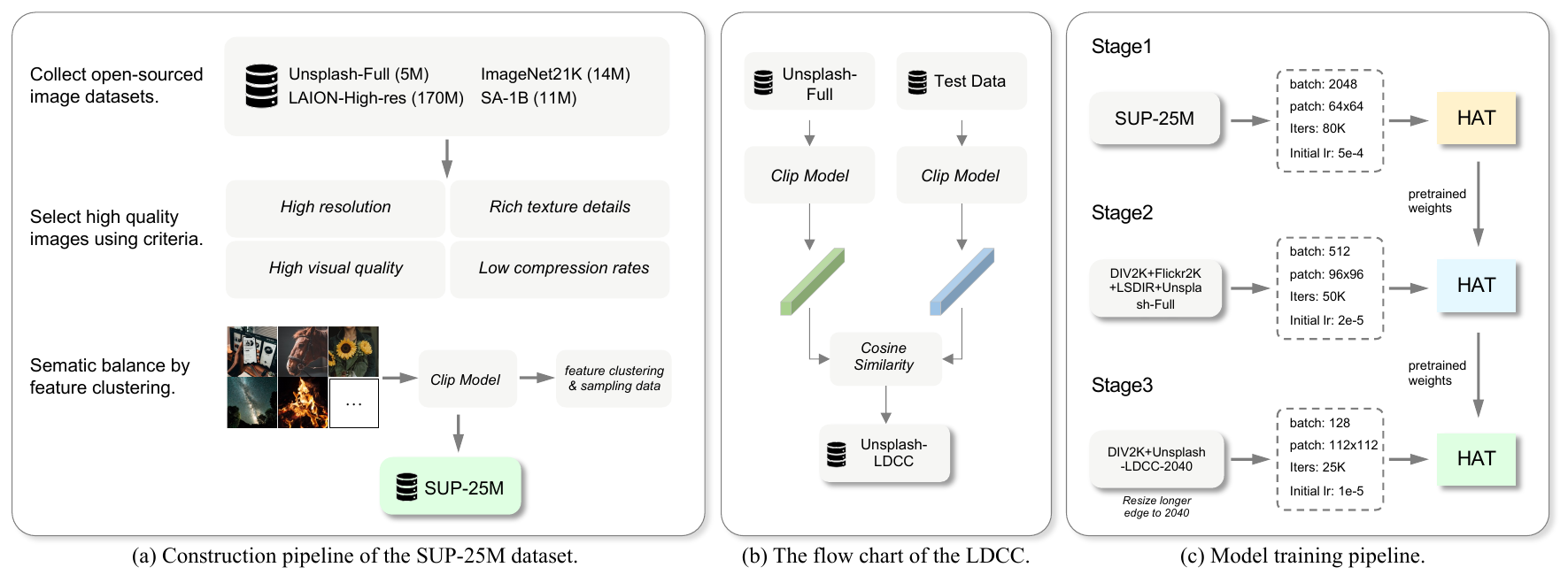}
    \caption{\textbf{Team SUPSR}}
    \label{fig:team2_01}
    \vspace{-4.mm}
\end{figure*}

\vspace{-5.mm}
\paragraph{Implementation Details.}
The dataset utilized for training comprises of DIV2K and LSDIR. During each training batch, 64 HR RGB patches are cropped, measuring $256$$\times$$256$, and subjected to random flipping and rotation. The learning rate is initialized at $5$$\times$$10^{-4}$ and undergoes a halving process every $2$$\times$$10^5$ iterations. The network undergoes training for a total of $10^6$ iterations, with the L1 loss function being minimized through the utilization of the Adam optimizer~\cite{kingma2014adam}. the authors repeated the aforementioned training settings four times after loading the trained weights. Subsequently, fine-tuning is executed using the L1 and L2 loss functions, with an initial learning rate of $1$$\times$$10^{-5}$ for $5$$\times$$10^5$ iterations, and HR patch size of 512. the authors conducted finetuning on four models utilizing both L1 and L2 losses, and employed batch sizes of 64 and 128. Finally, the authors integrated these four models to obtain the final model.

\textbf{K-means Based Fusion Strategy}: Given a set of $N$ super-resolution models ${M_1, M_2, \ldots, M_N}$, the authors first obtain their respective output images ${I_1, I_2, \ldots, I_N}$ for a given low-resolution input image. Each output image $I_i$ is then flattened into a one-dimensional vector $V_i$.

These vectors ${V_1, V_2, \ldots, V_N}$ serve as the input data for the K-means clustering algorithm. the authors perform K-means clustering to partition the vectors into $K$ disjoint clusters $C_1, C_2, \ldots, C_K$, where $K$ is a tunable hyperparameter for optimizing performance. The K-means algorithm aims to minimize the within-cluster sum of squared distances from each vector to its assigned cluster centroid.

For each cluster $C_k$, the authors calculate its cluster centroid $c_k$ as the mean of all vectors belonging to that cluster. This yields $K$ representative vectors ${c_1, c_2, \ldots, c_K}$, corresponding to the $K$ clusters. the authors then assign a weight $w_k$ to each cluster centroid $c_k$, proportional to the number of vectors belonging to cluster $C_k$. The rationale behind this weighting scheme is that clusters containing more model outputs should contribute more significantly to the final fused output.

The fused output image $I_f$ is obtained by computing the weighted sum of the cluster centroids:
\begin{equation}
I_f = \sum_{k=1}^{K} w_k \cdot c_k.
\end{equation}

Finally, the weighted sum vector $I_f$ is reshaped back into a two-dimensional image, representing the optimally fused output from the ensemble of super-resolution models.

The proposed K-means based fusion method effectively captures the diversity among the model outputs while mitigating potential outliers or poor predictions from individual models. By intelligently combining the outputs based on their inherent similarities, as identified by the clustering process, their approach aims to produce a superior fused output that leverages the collective strengths of ensemble.

\subsection{SUPSR}
\paragraph{Description.} Recently, many tasks have obtained astonishing improvements from scaling, such as SAM~\cite{SA1B2023}, Diffusion models~\cite{rombach2022high} and RAM~\cite{DBLP:journals/corr/abs-2306-03514}. Inspired by this, the authors make efforts to scale up training dataset to build a large-scale and intelligent restoration model. Specifically, they adopt the cutting-edge model, HAT~\cite{chen2022activating}, as the base model. For the training data, the team firstly collect a large amount open-sourced image datasets. Then a similar data selection strategy to LSDIR~\cite{li2023lsdir} and HQ-50K~\cite{HQ-50K} is employed to construct a dataset of 25 million high-quality images. Furthermore, self-ensemble is utilized to enhance the robustness and generalization.

\vspace{2.mm}
\textbf{Dataset.}
Inspired by the high-quality data construction schemes of DIV2K~\cite{timofte2017ntire}, LSDIR~\cite{li2023lsdir}, and HQ-50K~\cite{HQ-50K}, the authors propose a progressive high-quality data selection scheme. Subsequently, a dataset of \textit{25 million} (\textbf{SUP-25M}) high-quality images is constructed for model training.
Following pipeline for selecting high-quality images:
\vspace{2.mm}
\begin{itemize}
\vspace{.5mm}
\item \textit{High resolution:} Images with a resolution lower than $384$$\times$$384$ are removed.

\vspace{.5mm}
\item \textit{Compression rates:}
Images with bits per pixel (bpp) metric lower than 4 are removed in the pre-training stage and 12 in the fine-tuning stage.

\vspace{.5mm}
\item \textit{High visual quality:}
Image visual quality is measured by the non-reference image quality assessment (IQA) scores such as QPT~\cite{KVQ,Zoom-VQA}. The criterion was set to ensure that \( S_{iqa}>60 \) (within the range of 0 to 100) to achieve high-quality visual images.

\vspace{.5mm}
\item \textit{Rich texture details:} The ratio of power spectrum of the high-frequency components from~\cite{HQ-50K} was used, as well as blur and flat region detection from LSDIR~\cite{li2023lsdir}, as an indicator of texture selection metrics.

\vspace{.5mm}
\item \textit{Semantic balance:} A semantic selection based on Clip~\cite{radford2021learning} features was conducted to balance the richness of different scene categories in the dataset. 

\vspace{.5mm}
\item \textit{Specific domain:} In particular, greater use was made of the Unsplash~\cite{UnsplashData} full dataset during the model fine-tuning phase. A strategy consistent with DIV2K~\cite{timofte2017ntire} was followed, wherein the longer edge was resized to 2040, leading to the suppression of noise and compression. This adjustment allowed the characteristics of the Unsplash-Full dataset to better match those of the DIV2K image domain, thereby reducing the domain gap between training and testing.
\end{itemize}

\begin{figure}[t]
    \centering
    \includegraphics[width=\columnwidth]{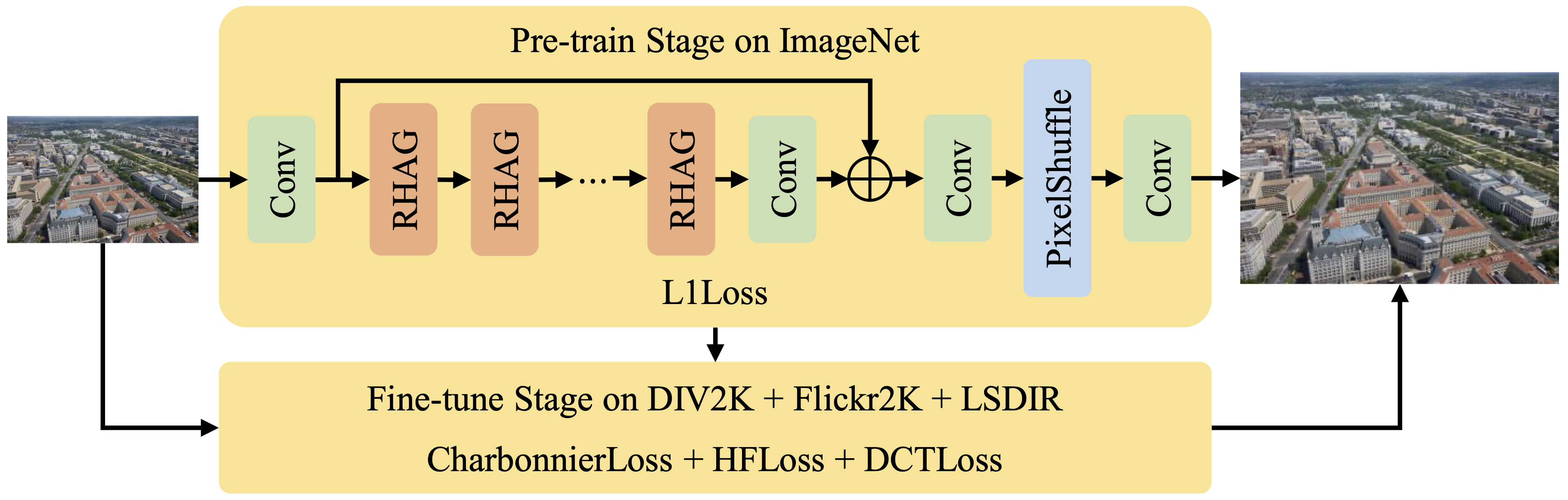}
    \caption{\textbf{Team UCAS-SCST}}
    \label{fig:team3_01}
\end{figure}

\vspace{4.mm}
\textbf{Methodology.}
The team adopted HAT~\cite{chen2022activating} as the base network model while the primary focus is on scaling up high-quality training data. HAT integrates channel attention and window-based self-attention, and currently holds the position as the leading model in classical image super-resolution. The overall architecture of the HAT is illustrated in \cref{fig:team2_01}, comprising three main components: shallow feature extraction, deep feature extraction, and image reconstruction modules. 

\vspace{-4.mm}
\paragraph{Implementation Details.}
The training process is divided into three stages. In the initial stage, the SUP-25M dataset is utilized with a patch size of $64$ and a batch size of $2048$, totaling $80K$ iterations with an initial learning rate of $5$$\times$$10^{-4}$. Stage two, built upon the previous checkpoint, undergoes further optimization for $50K$ iterations with a batch size of $512$ and a crop size of $96$, encompassing DIV2K, Flickr2K, LSDIR, Unsplash-Full, and HQ50K datasets. The learning rate is adjusted to $2$$\times$$10^{-5}$. In the final stage, the batch size reduces to $128$, crop size increases to $112$, and fine-tuning over $25K$ iterations is conducted on the DIV2K and Unsplash-Full datasets, selected with the LDCC strategy~\cite{zhang2023ntire_sr}, ensuring alignment between training and testing datasets.
The $L_1$ loss, Adam optimizer ($\beta_1$$=$$0.9$, $\beta_2$$=$$0.99$), and multi-step learning rate decay method are used during all three training phases. Also, data augmentation is performed on the training data through the flip and random rotation. 
In the testing phase, self ensemble, including flipping and rotating are used to enhance the final performance.
The total parameters of the team's network is 20.77M and FLOPs is 1037G on $256$$\times$$256$.

\begin{figure}[t]
    \centering
    \includegraphics[width=\columnwidth]{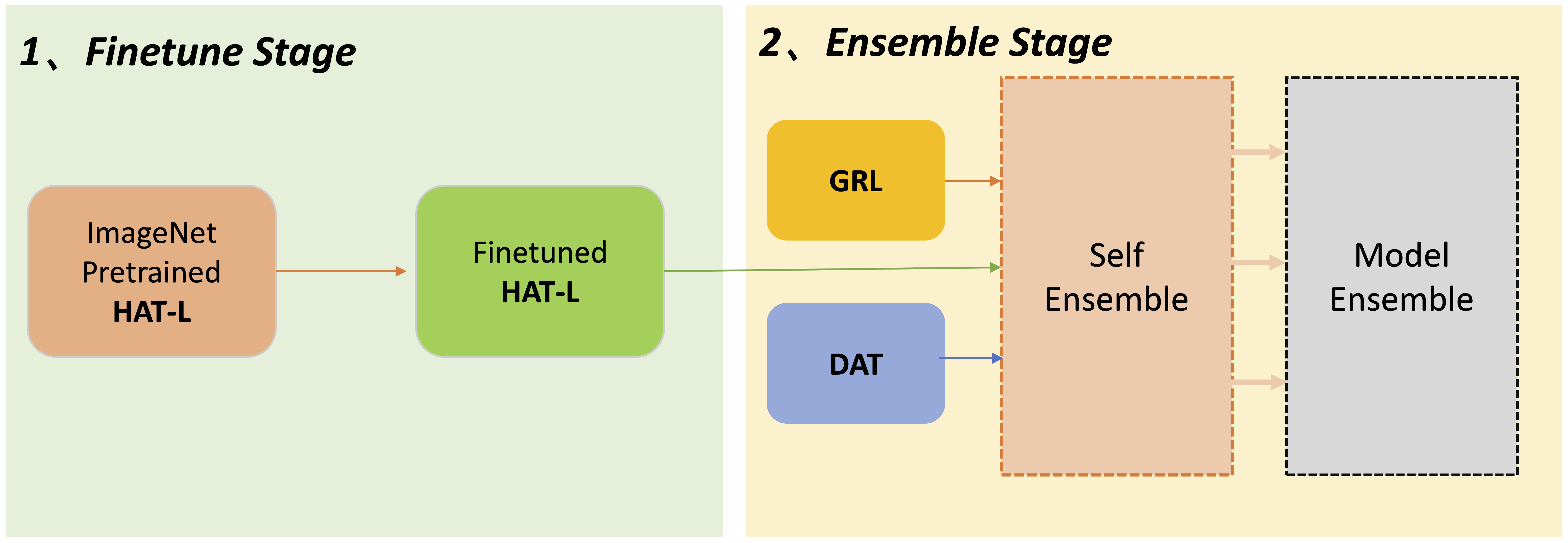}
    \caption{\textbf{Team SYSU-SR}}
    \label{fig:team04_01}
    \vspace{-2.mm}
\end{figure}

\vspace{-2.mm}
\subsection{UCAS-SCST}
\vspace{4.mm}
\paragraph{Description.}
Transformer-based methods have achieved impressive success in image restoration and enhancement tasks. Inspired by HAT\cite{chen2023activating}, which is an effective Transformer image super-resolution model, the authors propose a $\times$$4$ image super-resolution Transformer model, namely High Frequency Transformer (HFT). As shown in \cref{fig:team3_01}, the training process of the HFT model is divided into two stages, the pre-train stage and the fine-tune stage. We take the HAT-L \cite{chen2023activating} pre-trained model as a baseline and further optimize it in the fine-tune stage. Based on HAT-L\cite{chen2023activating}, they incorporate the high frequency loss to reduce the compression artifacts. Specifically, they constrain the residuals after Gaussian blurring and images in frequency domain obtained by Discrete Cosine Transform (DCT). First, the loss of high-frequency (HF) information is formulated as:
\begin{equation}
 L_{HF} = \big\| (I_{HR} - \text{B}(I_{HR})) - (I_{SR} - \text{B}(I_{SR})) \big\| _1,
\end{equation}
where $I_{HR}$ and $I_{SR}$ indicate the ground-truth HR image and reconstructed SR image, B($\cdot$) denotes a 5 $\times$ 5 kernel gaussian blur operation. Second, we calculate the loss in the DCT frequency domain as follows:
\begin{equation}
 L_{DCT} = \big\| \text{DCT}(I_{HR}) - \text{DCT}(I_{SR}) \big\| _1,
\end{equation}
where DCT($\cdot$) represents the Discrete Cosine Transform process. Moreover, the loss function of HFT is defined as: \begin{equation}
 L =  L_{Charbonnier} + \alpha L_{HF} + \beta L_{DCT},
\end{equation}
where the weight parameters $\alpha$ and $\beta$ are preset to 1 and 0.001. Through the optimization of the above loss function, HFT is more robust and reconstructs more details.

\vspace{-2.mm}
\paragraph{Implementation Details.} The DIV2K and LSDIR datasets, supplemented by Flickr2K and ImageNet datasets, serve as training data. Employing rotating and flipping strategies enhances these images. To expedite training, a portion of the network is initialized with the pre-trained HAT model on the ImageNet dataset. Utilizing the Adam optimizer ($\beta_1$$=$$0.9$, $\beta_2$$=$$0.99$), the authors conduct training for 300 iterations across 8 NVIDIA A100 GPUs, with a batch size of 8 and a patch size of 64$\times$64. The HFT model comprises 40.85 million parameters with an input size of 64$\times$64$\times$3. During inference, the authors implement a self-ensemble strategy to enhance performance.

\begin{figure*}[t]
    \centering
    \includegraphics[width=\textwidth]{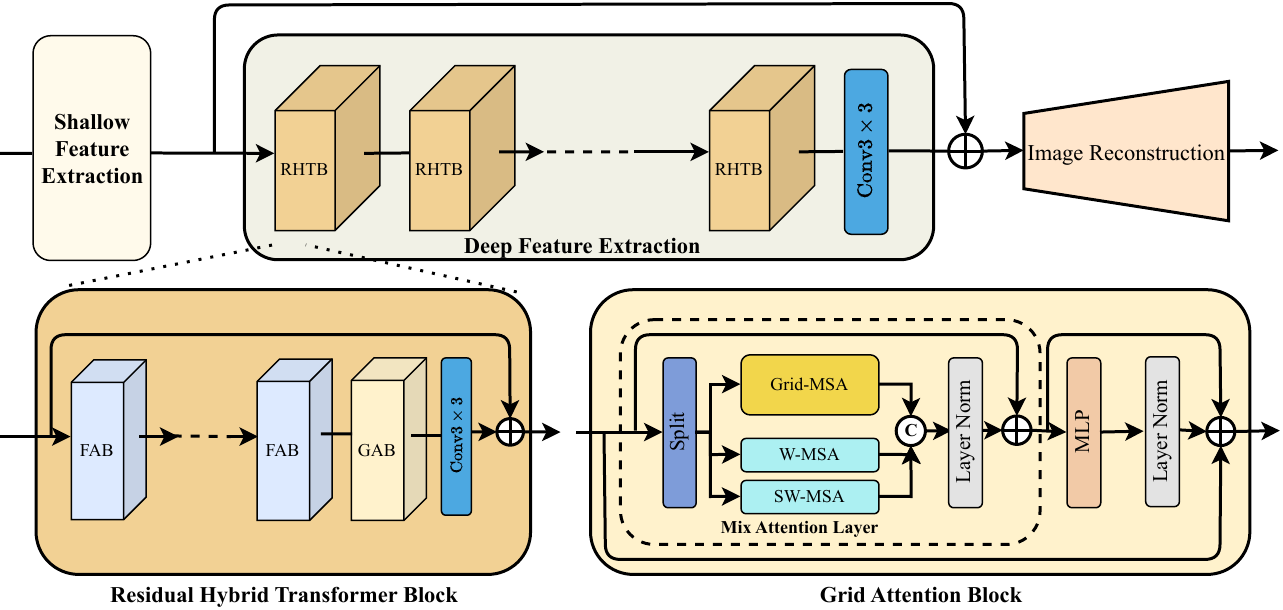}
    \caption{\textbf{Team Jasmine}}
    \label{fig:team05_01}
    \vspace{-4.mm}
\end{figure*}

\subsection{SYSU-SR}
\paragraph{Description.} The whole model workflow proposed by the team is shown in \cref{fig:team04_01}. Inspired by the excellent performance of large pre-trained model on the low-level computer vision tasks, the authors chose the HAT-L~\cite{chen2023activating} pre-trained model as their main structure and the authors proposed enhancement in the two phases of training and testing, respectively.

\textit{Training:} In order to further improve the performance of the pre-trained model, the team finetunes the pre-trained model from two main aspects: the loss function and the training strategy.
The loss function uses L1loss and Gradient-Weighted (GW) loss~\cite{wei2020component}, which constrains image's local structure and texture to generate more accurate details, which is efficient for the super-resolution task.
The training strategy uses progressive training, and some works ~\cite{liang2021swinir,zamir2021restormer} demonstrates its effectiveness.

\textit{Testing:} In order to further reduce the model's prediction bias on the super-resolution, the team fused two ensemble learning strategies, self-ensemble as well as model-ensemble, to enhance the model's test performance.
The authors first employ the self-ensemble approach~\cite{lim2017enhanced} to enhance all candidate models. Secondly, the model-ensemble approach proposed by ZZPM team~\cite{zhang2023ntire_sr}is applied to all enhanced models, where the average of the outputs of different models is calculated, and then the weights of the ensemble are assigned based on the MSE value between each model and the average. The experiments show that the fusion of two ensemble strategies can achieve higher performance than each single strategy.

\paragraph{Implementation Details.}
In the model finetuning phase, the training data contains DIV2K, Flickr2K and LSDIR and the training loss is $\mathcal{L}_{total}=\alpha\cdot\mathcal{L}_{1}+\beta\cdot\mathcal{L}_{GW}$ where $\alpha$ and $\beta$ are weights assigned to the two losses. the authors set $\alpha$$=$$1$, $\beta$$=$$3$ as their training setup and all  experiments is conducted on 8 NVIDIA A100 GPUs using Adam optimizer. At the beginning of progressive training, the patch size is 64 and batchsize is 32, keeping the learning rate as $1$$\times$$10^{-5}$ for 125$k$ iterations. Finally setting the patch size as 128, batchsize as 16 with the learning rate as $5$$\times$$10^{-6}$ for 60$k$ iterations.

In the model testing phase, the pre-trained GRL~\cite{li2023efficient}, DAT~\cite{chen2023dual} models and the finetuned HAT-L model are selected for fusion of outputs to obtain higher performance.

\begin{figure*}[t]
    \centering
    \includegraphics[width=\textwidth]{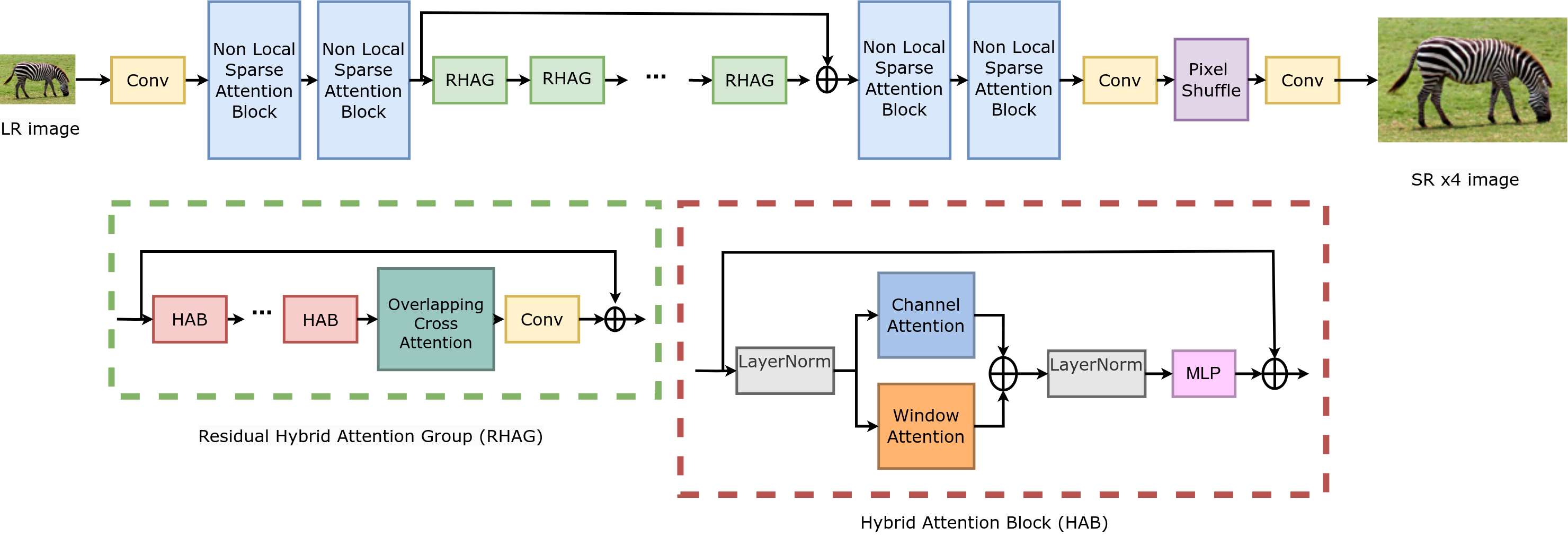}
    \caption{\textbf{Team mandalinadagi.} Main architecture.}
    \label{fig:team7_01}
    \vspace{-2.mm}
\end{figure*}

\begin{figure}[t]
    \centering
    \includegraphics[width=\columnwidth]{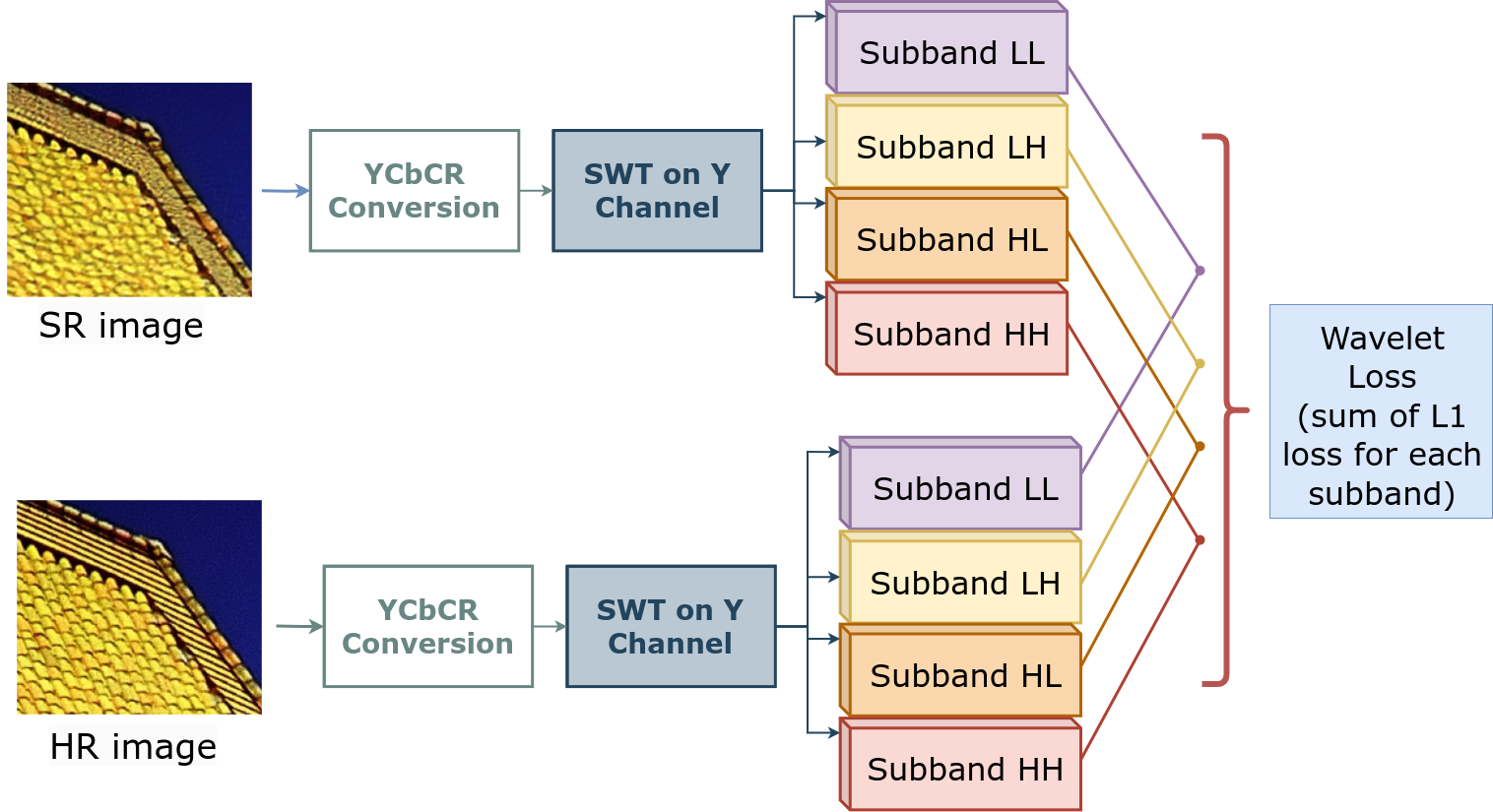}
    \caption{\textbf{Team mandalinadagi.} Wavelet loss.}
    \label{fig:team7_02}
    \vspace{-2.mm}
\end{figure}

\subsection{Jasmine}
\paragraph{Description.} As shown in \cref{fig:team05_01}, the team proposes an image super-resolution Transformer with cross-scale attention and Fused Conv\cite{tan2019efficientnet} for solving super-resolution tasks.

In cross-scale attention, the team introduces Grid Attention\cite{zhang2022accurate} and add (S)W-MSA. the feature maps are split by channel and put into Grid Attention and (S)W-MSA, respectively. this approach allows for the extraction of global and local features and the modeling of multi-scale information. Improve the efficiency of utilizing duplicate texture features.

In addition, the authors introduce Fused Conv to enhance the feature representation. The Fused Conv block with an expansion rate of 6 and a shrink rate of 4 is added at the beginning to each successive (S)W-MAS block. This combination of CNN and Transformer helps extract finer details.

\paragraph{Implementation Details.} The model is trained on the combination of the trainsets of DIV2K, Flickr2K, and LSDIR. The training process is divided into two stages. In the first stage training on the LSDIR training set, the model is trained on 64$\times$64 randomly cropped images with The Adam optimizer. The batch size, initial learning rate, and total iterations are 32, $2$$\times$$10^4$ and 800K, respectively. The learning rate is halved at 300K, 500K, 650K, 700K, and 750K iterations. Training on the DF2K training set in the second stage, the model is fine-tuned on 64$\times$64 images with 250K iterations. The learning rate is halved at 125K, 200K, 230K, and 240K iterations.

\subsection{ACVLAB}
\paragraph{Description.} ACVLAB proposes a solution leveraging HAT \cite{chen2023activating} and a self-ensemble fusion method for enhancement. They introduce Dense-residual-connected Transformer (DRCT) to enhance the receptive fields by integrating multi-level residual and dense connections. DRCT comprises three main components: (1) shallow feature extraction using a single 3$\times$3 convolutional kernel, (2) deep feature extraction employing 12 Residual Deep feature extraction Group (RDG)   with Swin-Dense-Residual-Connected Blocks (SDRCB) for learning long-range dependencies, and (3) image reconstruction through 3$\times$3 convolutional kernels and upsampling.

\paragraph{Implementation Details.} ACVLAB trained their models using the LSDIR dataset \cite{li2023lsdir} and the DIV2K dataset \cite{timofte2017ntire}, structured into two phases. They employed the Adam optimizer with parameters $\beta_{1}$$=$$0.9$ and $\beta_{2}$$=$$0.999$, running for 800,000 iterations per phase. The learning rate of $2$$\times$$10^{-4}$ was reduced by half at iterations [300k, 500k, 650k, 700k, 750k] using a multi-step scheduler. After convergence, three models were generated without weight decay. High-resolution (HR) patches of 256$\times$256 pixels were extracted for data preparation, augmented with random flips and rotations. L1 loss was utilized with a batch size of 16 in the first phase, transitioning to MSE loss for the second phase. Testing-time augmentation (TTA) was implemented through self-ensembling, incorporating rotation, horizontal, and vertical flipping. Additionally, model ensembling was performed for result fusion by HAT and the proposed DRCT. Implementation was done using PyTorch 1.13.1, trained on two NVIDIA GeForce RTX 3090 GPUs.

\begin{figure*}[t]
    \centering
    \includegraphics[width=\textwidth]{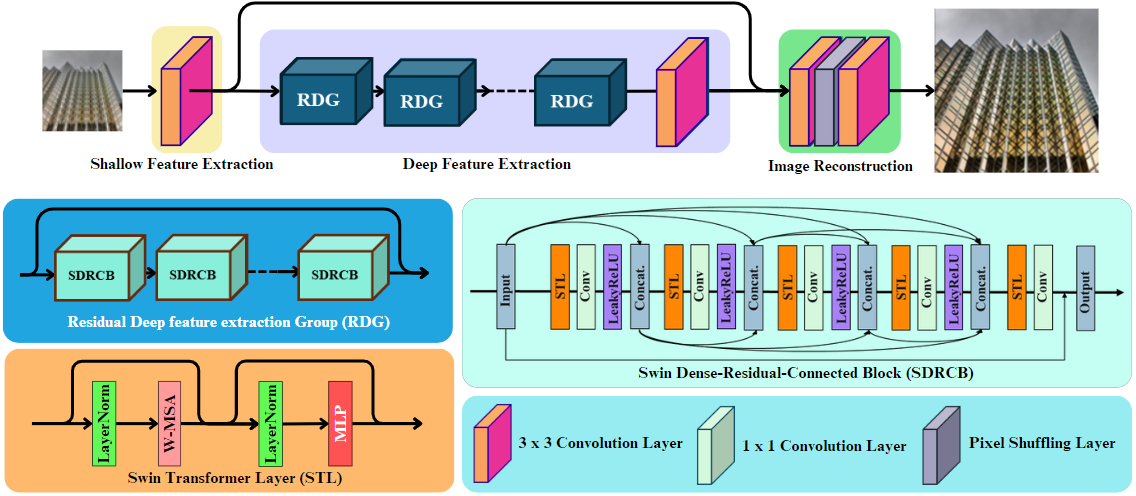}
    \caption{\textbf{Team ACVLAB.} Overall architecture.}
    \label{fig:team6_01}
    \vspace{-2.mm}
\end{figure*}

\begin{figure}[t]
    \centering
    \includegraphics[width=\columnwidth]{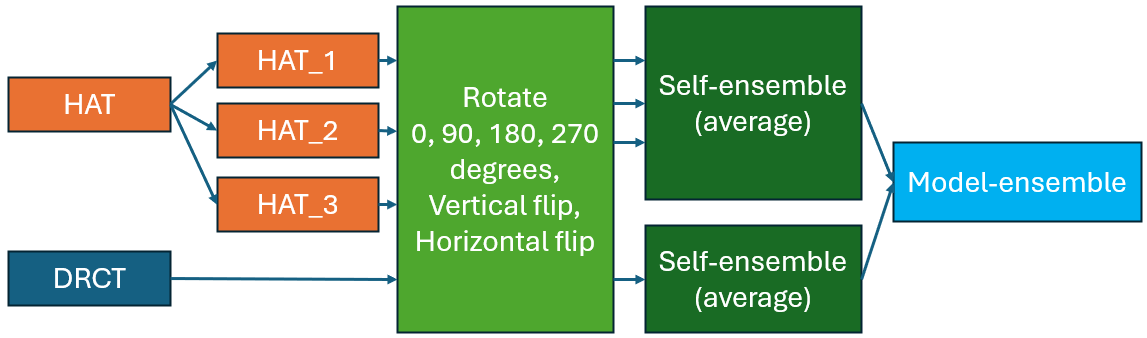}
    \caption{\textbf{Team ACVLAB.} Inference pipeline.}
    \label{fig:team6_02}
    \vspace{-4.mm}
\end{figure}

\subsection{mandalinadagi}
\paragraph{Description.} The proposed approach \cite{korkmaz2024wavelettention} builds upon the HAT architecture \cite{chen2023activating} as a baseline and enhances it by sandwiching it between non-local attention blocks \cite{mei2021imageNLSA} depicted in \cref{fig:team7_01}. The baseline HAT architecture utilized Residual-in-Residual approach \cite{wang2018esrgan} and developed hybrid attention block (HAB) similar to the standard Swin Transformer block \cite{liang2021swinir} to enhance performance. In addition, the Non-Local Sparse Attention (NLSA) module combines several techniques to enhance efficiency and global modeling in attention mechanisms. Specifically, NLSA uses Spherical Locality Sensitive Hashing method to divide input features into buckets to calculate attention. This module incorporates the strengths of Non-Local Attention, which allows for global modeling and capturing long-range dependencies in data. Additionally, it leverages the advantages of sparsity and hashing, which lead to high computational efficiency. Therefore, by combining HAT architecture with the NLSA module, the authors aim to achieve effective attention mechanisms with reduced computational costs, making it suitable for various applications in deep learning models especially for SR. Additionally, this modified architecture is trained using wavelet losses along with the $L_1$ RGB loss. In details, the Stationary Wavelet Transform (SWT) is a technique that enables the multi-scale decomposition of images \cite{wavelet_doc}. This process results in one low-frequency (LF) subband called LL and several high-frequency (HF) subbands called LH, HL, and HH. The number of HF subbands is determined by the decomposition level of the LL subband, and each HF subband contains detailed information in horizontal, vertical, or diagonal directions. Since SWT inherently combines scale/frequency information with spatial location, making it particularly suitable for tasks where preserving spatial details across different scales is essential, such as in SR applications. This combination of SWT loss with the proposed transformer model aims to improve the overall performance of the baseline HAT model \cite{chen2023activating} by incorporating non-local attention mechanisms and leveraging wavelet losses to enhance image quality during training.

The wavelet transform (WT) is known for its efficiency and intuitive ability to represent and store multi-resolution images effectively \cite{mallat1996wavelets}. This means that it can capture both contextual and textural information of an image across different levels of detail. The understanding of how WT operates and its capacity to handle varying levels of image details inspired us to integrate wavelet-domain losses into a transformer-based super-resolution system. In other words, wavelet-domain subbands have capability of handling different aspects of information encoded in images that enables it a promising addition to enhance the performance of a super-resolution system based on transformers. Hence, rather than employing the typical RGB-domain fidelity loss seen in conventional transformer methods, the authors introduce the SWT-domain fidelity loss denoted as $L_{SWT}$ along with corresponding tuning parameter is illustrated in \cref{fig:team7_02}. The $l_1$ fidelity loss is calculated between the SWT subbands of the generated images $x$ and the ground truth (GT) image $y$, and then averaged over a mini-batch size represented by $\mathbb{E}[.]$. This formulation allows for a more nuanced optimization process that can enhance the overall quality of the super-resolved images produced by the image transformer models:
\begin{multline}  \label{eq:swt_fidelity}
 L_{SWT} = \mathbb{E} \bigl[ \sum_{j} \lambda_{j} \big\| {SWT(G(x))}_{j} - {SWT(y)}_{j} \big\| _1  \bigr],
\end{multline} 
where $G$ denotes the proposed SR model and $\lambda_j$ are appropriate scaling factors to control the generated HF details. To properly scale each subband the authors set $\lambda_j$ to 0.05.

Then, the overall loss for the training is given by 
\begin{equation}
 L_{G} = L_{RGB} + L_{SWT},
\end{equation} 
where $L_{RGB}$ denotes the $l_1$ loss calculated on RGB domain, measuring pixel-wise errors in the image space.

\paragraph{Implementation Details.}
The authors configured the chunk size for the non-local sparse attention as 144 and added 2 consecutive NLSA \cite{mei2021imageNLSA} layers before and after the HAT \cite{chen2023activating} architecture. The authors utilized pre-trained HAT-L as in provided configuration. Hence the embedding dimension is set to 180 and patch embedding is set to 4. The total parameter number of proposed method is 41.27M. During training, the authors randomly crop 64$\times$64 patches from the LR images from LSDIR \cite{li2023lsdir} and DIV2K \cite{timofte2017ntire} datasets to form a mini-batch of 8 images. The training images are further augmented via horizontal flipping and random rotation of 90, 180, and 270 degrees. the authors optimize the model by ADAM optimizer \cite{kingma2014adam} with default parameters. The learning rate is set to $4e^5$ and reduced by half after 125k, 200k and 240k iterations. The final model is obtained after 250k iterations. Their model is implemented with PyTorch and trained on Nvdia A40 GPUs with a total compute demand of 81.33 GFLOPS.

\begin{figure}[t]
    \centering
    \includegraphics[width=\columnwidth]{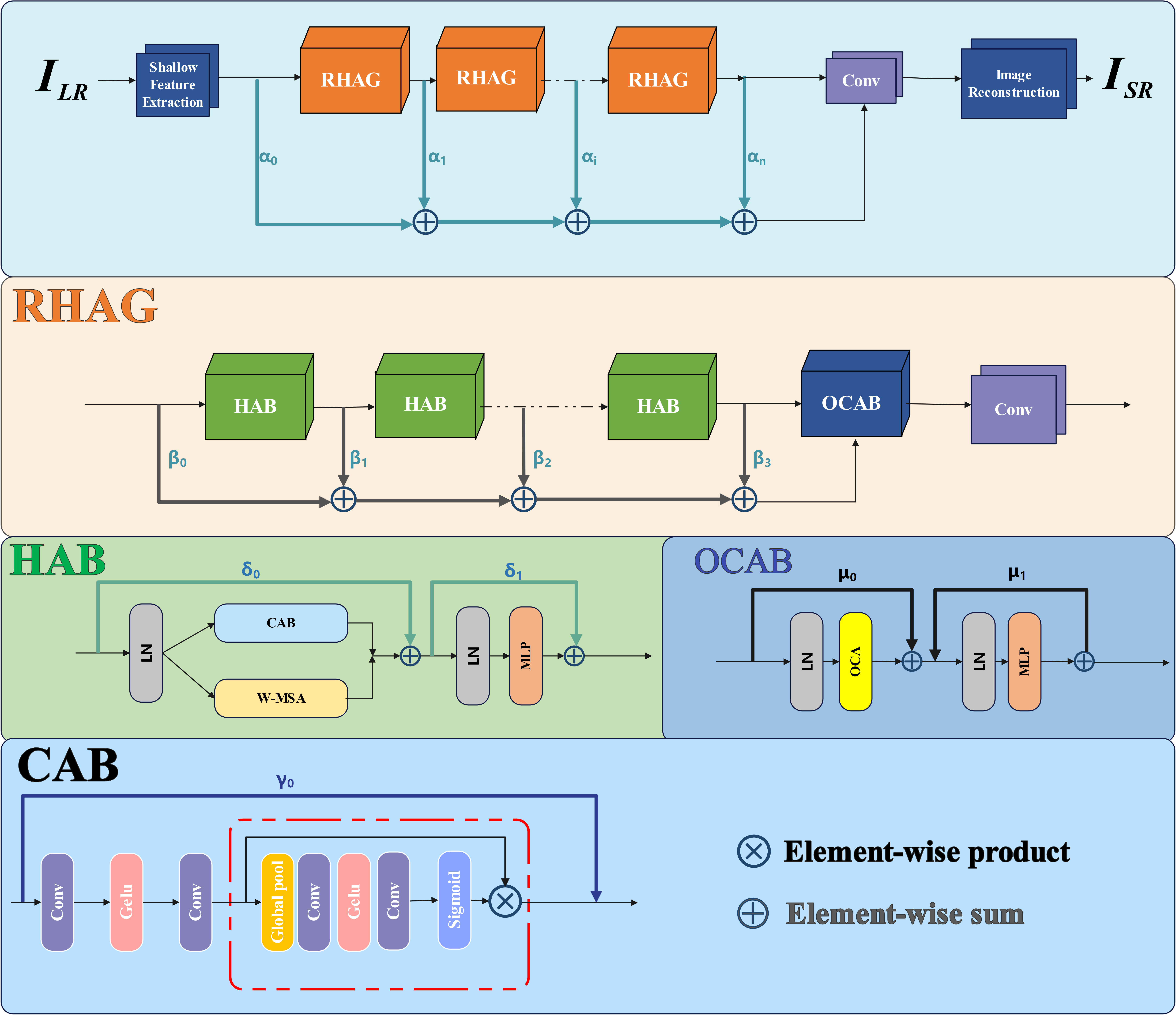}
    \caption{\textbf{Team SKDADDY}}
    \label{fig:team08_01}
    \vspace{-2.mm}
\end{figure}

\subsection{SKDADDYS}
\paragraph{Description.} The network architecture of EHAT in ~\cref{fig:team08_01}, has been improved based on the HAT network. In the EHAT network, we introduced a Learnable Branch Weight Coefficient (LBWC).Compared to the HAT network, which simply sets the branch weights to 0 or 1, LBWC employs a learning mechanism that accurately identifies the importance of each branch and dynamically adjusts the weights accordingly. Consequently, the model can more effectively capture key features in the data. Ideally, LBWC assigns larger weight coefficients to branches with more significant features, thereby more effectively influencing the overall network performance.

\vspace{-4.mm}
\paragraph{Implementation Details.}
\vspace{4.mm}

\begin{itemize}
\item \textbf{Track1}: Firstly, the authors utilized the HAT-L\_SRx4\_ImageNet-pretrain.pth model from the official HAT website as their pre-trained model. Further, the authors employed the original dataset of 800 images from DIV2K and 10,000 images from LSDIR as training set. Subsequently, the authors improved the HAT model by introducing the BranchAttentionModule and fine-tuned the model with a batch size of 2. Training was conducted for 500 iterations on an NVIDIA GeForce RTX 3090 GPU. Additionally, the model was optimized using Adam with $\beta_{1}$$=$$0.9$ and $\beta_{2}$$=$$0.99$, and default weight decay of zero. The initial learning rate was set to $10^{-5}$ and preliminary training was performed using L1 loss.

\item \textbf{Track2}: Following Track1, the authors selected the best weight (measured by PSNR) from Track1 as the pre-trained model for this round. Furthermore, the authors set the learning rate to $10^{-8}$ and continued training the model.

\item \textbf{Track3}: Finally, utilizing the pre-trained model from Track2, the authors sequentially adjusted the learning rate from $10^{-5}$ to $10^{-8}$ and replaced the L1 loss function with MSELoss for fine-tuning. Subsequently, the data was sent into the network for training to get the final results.

The final experimental results are shown in \textbf{figure 3}. The results show that EHAT has a significant improvement compared with RealESRGAN, HAT and HATGAN.
\end{itemize}

\begin{figure}[t]
    \centering
    \includegraphics[width=\columnwidth]{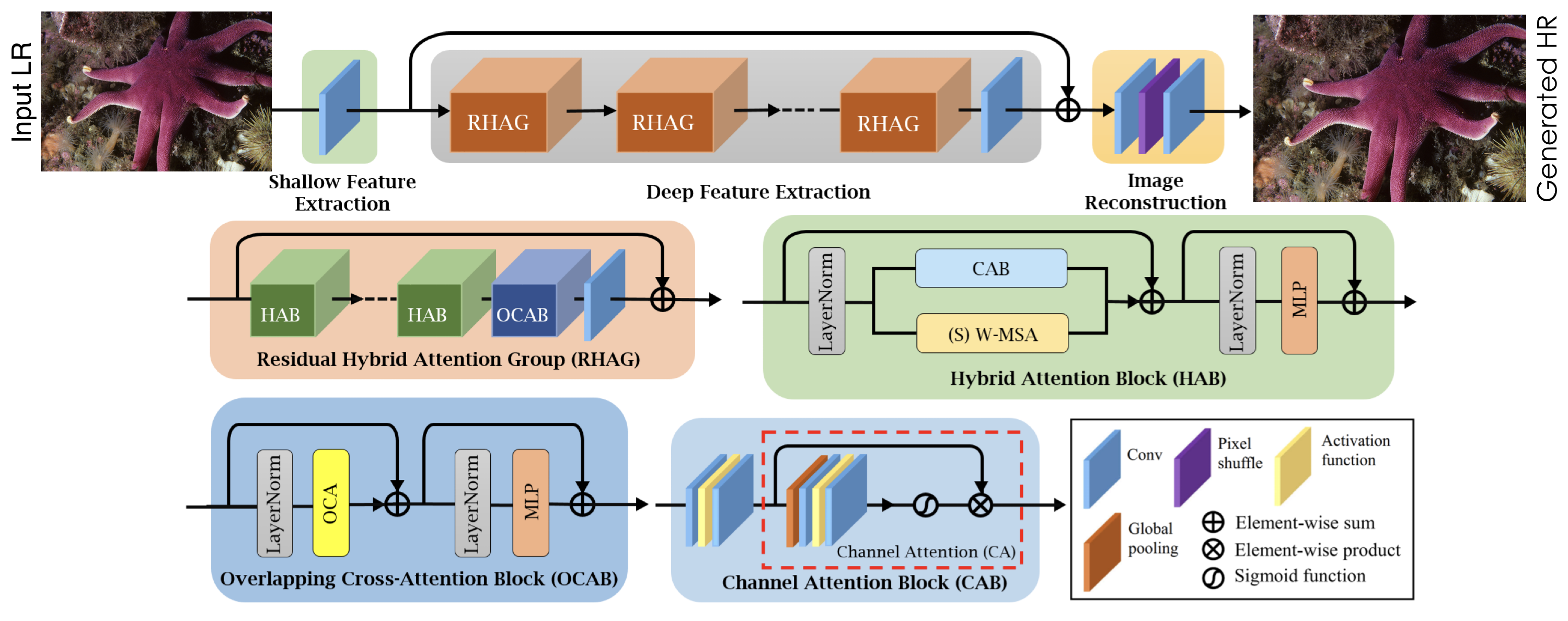}
    \caption{\textbf{Team KLETech-CEVI}}
    \label{fig:team09_1}
\end{figure}

\begin{figure*}[t]
  \centering
  \footnotesize
    \begin{subfigure}[t]{\textwidth}
      \centering
      \includegraphics[width=\textwidth]{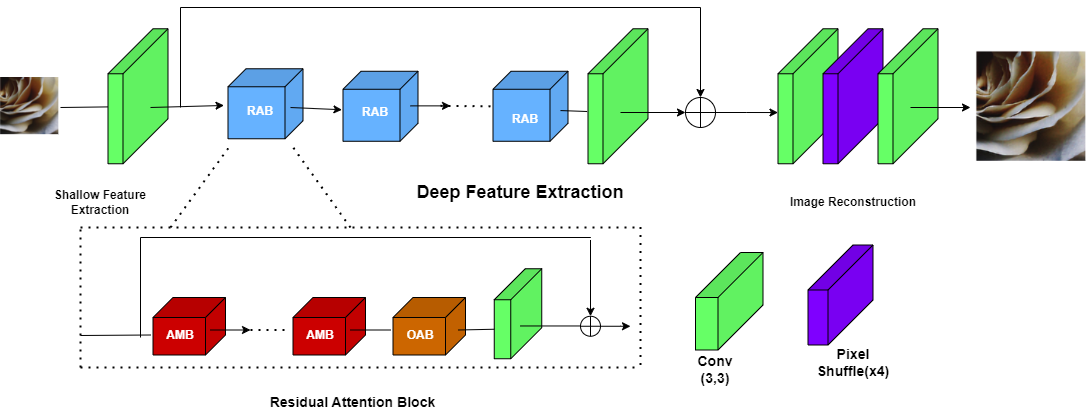}
      \caption{}
      \label{fig:team10_01}
    \end{subfigure}
    \hfill
    \begin{subfigure}[t]{\columnwidth}
      \centering
      \includegraphics[width=\columnwidth]{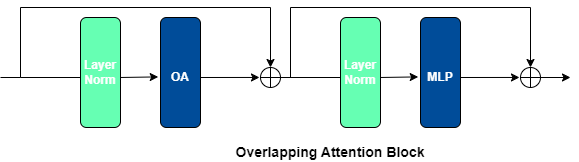}
      \caption{}
      \label{fig:team10_02}
    \end{subfigure}
    \hfill
    \begin{subfigure}[t]{\columnwidth}
      \centering
      \includegraphics[width=\columnwidth]{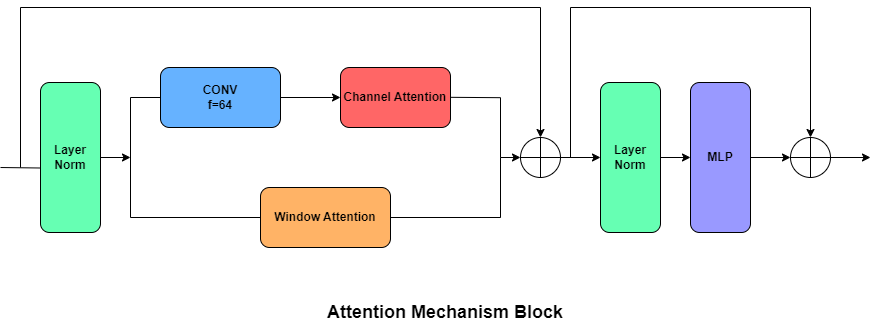}
      \caption{}
      \label{fig:team10_03}
    \end{subfigure}
  \caption{\textbf{Team SVNIT-NTNU}}
\end{figure*}

\subsection{KLETech-CEVI-Lowlight-Hypnotise}
\paragraph{Description.} Image super-resolution techniques like \cite{5557890}, \cite{patil2018evidence}, \cite{chen2024mffn}, \cite{al2024single} aim at improving the resolution of images. The network architecture inspired from \cite{chen2023activating} for the image super-resolution task the authors proposed SR4X, as illustrated in \cref{fig:team09_1}. SR4X comprises of three main components: initial feature extraction, deep feature extraction, and image reconstruction with combinational loss function. This design has been employed in prior studies. For a given low-resolution (LR) input $I_{LR}$$\in$$\mathbb{R}^{H \times W \times C}$, the process begins with a convolutional layer that extracts shallow features $F_0$$\in$$\mathbb{R}^{H \times W \times C}$, where \(C\) represents the channel number of the input and the intermediate feature. Subsequently, a sequence of residual hybrid attention groups (RHAG) and a single 3$\times$3 convolutional layer, denoted as \(HConv(\cdot)\), are employed for deep feature extraction.

Following this, a global residual connection is introduced to merge the shallow features \(F_0\) with the deep features $F_D$$\in$$\mathbb{R}^{H \times W \times C}$. The high-resolution output is then generated through a reconstruction module. As depicted in \cref{fig:team09_1}, each RHAG comprises multiple hybrid attention blocks (HAB), an overlapping cross-attention block (OCAB), and a 3$\times$3 convolutional layer with a residual connection. For the reconstruction phase, the pixel-shuffle method is employed to upscale the combined feature. The network parameters are optimized using a weighted combination of L1 and VGG-19 perceptual loss \cite{ledig2017photo}:
\begin{equation}
    \mathcal{L}_{SR} = \alpha * \mathcal{L}_{VGG} + \beta * \mathcal{L}_{1},
    \label{eqn:ile}
\end{equation}
where, $\mathcal{L}_{VGG}$ is VGG-19 perceptual loss \cite{ledig2017photo}. $\alpha$ and $\beta$ are weights to the losses, and are set to $0.7$ and $0.5$ heuristically.

\vspace{-4.mm}
\paragraph{Implementation Details.}
\begin{itemize}
    \item Dataset used to train the model include LSDIR \cite{li2023lsdir}, DIV2K \cite{timofte2017ntire} datasets.
    \item Training of the proposed methodology is conducted using the dataset provided by NTIRE 2024 Efficient Super-Resolution Challenge. the authors train the model using Python and PyTorch frameworks, on patch resolution of 339*510, with a batch size of 8. the authors use Adam optimizer with $\beta_1$ set to $0.9$ and $\beta_2$ set to $0.999$. The authors train the model for 1000 epochs at a learning rate of $0.0002$.
    \item Testing description: During testing, the authors use full resolution images (339*510), on single RTX 3090 GPU. Average testing time for single image on full resolution is 0.9s on RTX 3090 GPU.
\end{itemize}

\begin{figure*}[t]
    \centering
    \includegraphics[width=\textwidth]{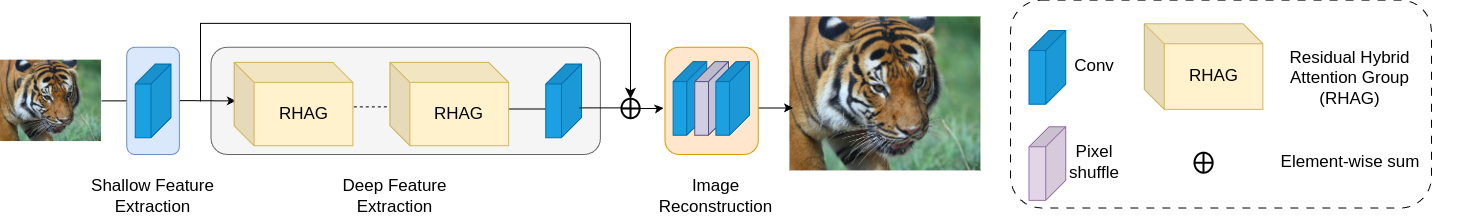}
    \caption{\textbf{Team ResoRevolution.}}
    \label{fig:team11_01}
    \vspace{-2.mm}
\end{figure*}

\subsection{SVNIT-NTNU}
\paragraph{Description.}  In order to design single image super-resolution, the authors use Transformer and CNN based network approach in the proposed solution. As shown in \cref{fig:team10_01}, the overall network consists of three parts, including shallow feature extraction, deep feature extraction and image reconstruction.
The \cref{fig:team10_01} depicts the proposed architecture for single image super-resolution for scaling factors of $\times4$. The LR image is applied as input to the network and it is passed to extract the salient features from it. The low frequency features are extracted with first layers that employ convolutional layer While high frequency features are extracted with deep feature extraction bock. The architecture uses the Exponential Linear Unit (ELU) activation function to improve learning performance at each layer in efficient manner. The global residual connection to fused shallow features and
deep features and then reconstruct the high-resolution result via a reconstruction module. The residual attention block preserves the high frequency details of the SR image by retaining salient features which is displayed in \cref{fig:team10_01}. Each residual attention block contains several attention mechanism block (AMB), an overlapping attention block (OAB) and a 3 × 3 convolution layer with a residual connection. The architecture of AMB and OAB is depicted in \cref{fig:team10_02} and \cref{fig:team10_03} respectively. The channel attention block are further used in AMB to perform adaptive re-scaling of features on per-channel basis. The pixel Shuffle is used to upscale the feature maps to the desired scaling factor (\ie, $\times4$) \cite{chen2023activating}.

\vspace{-2.mm}
\paragraph{Implementation Details.}
The authors use DIV2K and LSDIR dataset for training and keep the depth and width the same as SwinIR. Specifically, the RAB number and AMB number are both set to 8. The channel number is set to 64. The attention head number and window size are set to 6 and 16.The code is implemented using Pytorch library. The loss function is weighted combination of $l_1$, SSIM and Charbonnier loss with a learning rate of $1 \times 10^{-4}$ which is decayed by $1\times10^4$ iterations and the same is optimized using Adam optimizer. The model is trained up to $1 \times 10^5$ iterations with a batch size of $8$.

\begin{figure*}[t]
	\centering
	\includegraphics[width=\textwidth]{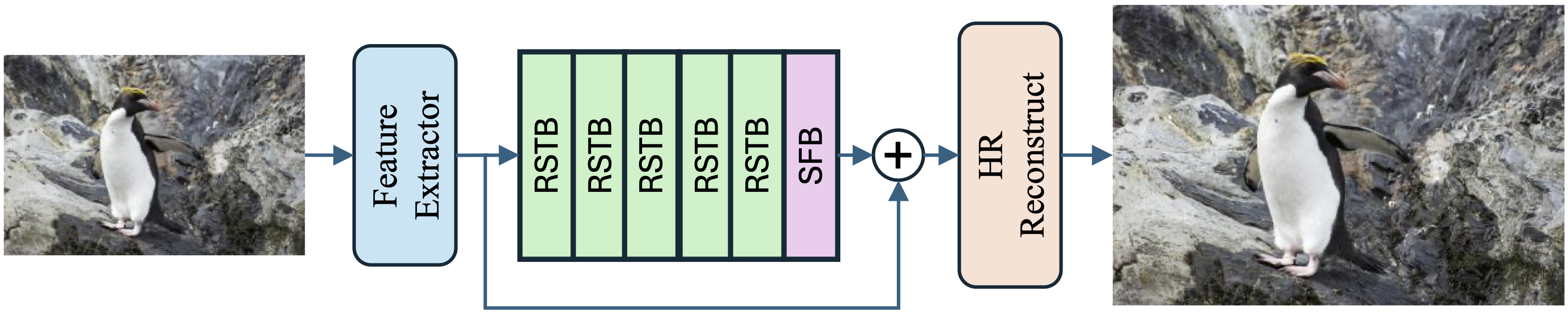}
	\caption{\textbf{Team BetterSR}}
	\label{fig:team12_01}
\end{figure*}

\vspace{4.mm}
\subsection{ResoRevolution}
\paragraph{Description.} The proposed methodology utilizes the Hybrid Attention Transformer (HAT) model  \cite{chen2023activating} as the foundation for addressing the task of image Super-Resolution ($\times$4). 
HAT's architecture is segmented into three primary components: shallow feature extraction, deep feature extraction, and image reconstruction. Initially, a convolution layer processes the low-resolution (LR) input to extract shallow features. Subsequently, these features undergo deep feature extraction via a sequence of residual hybrid attention groups (RHAGs) and an additional convolution layer. The deep features and shallow features are then merged through a global residual connection.
Hybrid Attention Block (HAB): At the heart of HAT is the Hybrid Attention Block (HAB), which synergizes channel attention with window-based multi-head self-attention (W-MSA). The inclusion of a Channel Attention Block (CAB) within the standard Transformer block, parallel to the W-MSA module, empowers the model to leverage global information for channel attention while maintaining robust local feature representation through self-attention. This dual attention mechanism facilitates more effective pixel activation, crucial for enhanced SR reconstruction.
Overlapping Cross-Attention Block (OCAB): To further augment feature interaction across different windows, HAT introduces the Overlapping Cross-Attention Block (OCAB). OCAB employs an overlapping window partitioning strategy to extend the receptive field beyond individual windows, enabling the model to integrate cross-window information more effectively. This approach not only addresses the limitations posed by non-overlapping windows but also significantly reduces blocking artifacts in the reconstructed images, leading to clearer and more coherent SR results. The model architecture is shown in \cref{fig:team11_01}.

\vspace{-2.mm}
\paragraph{Implementation Details.} Training Strategy: The training of the proposed model was executed in two principal stages:
Initial Training: The model was trained by combining the DF2K and Imagenet dataset. DF2K serves as the primary training dataset and is a blend of DIV2K and Flicker2K datasets, totaling 3450 images.
For pre-training, the HAT model leverages the full ImageNet dataset, consisting of approximately 1.28 million images. This large-scale dataset is instrumental in the same-task pre-training strategy adopted by HAT, aiming to exploit the potential of the model for further improvement in super-resolution performance. The ImageNet dataset, with its vast diversity in image content, serves as a robust foundation for pre-training, enabling the HAT model to learn generalizable features that are refined during the subsequent fine-tuning phase on the DF2K dataset.
The model was initially trained with a focus on learning from 64$\times$64 randomly cropped images. This stage utilized the Adam optimizer, with a batch size of 32, an initial learning rate of 0.0002, and a total of 800K iterations. The learning rate was methodically reduced at pre-defined checkpoints (300K, 500K, 650K, 700K, and 750K iterations) to facilitate optimal convergence.

Fine-tuning: In the subsequent stage, the model was fine-tuned on larger 128$\times$128 cropped images for an additional 200K iterations. This fine-tuning process is essential for the model to adapt its learned features to reconstruct higher-resolution images more accurately. the authors utilised DIV2K and LSDIR dataset in this stage. During the training phase, an L1 loss is used for SR, and an MSEloss is used for enhancement. The training of the aforementioned model is conducted on 4 Nvidia RTX 3090 GPUs, using the Adam optimizer and multistep learning rate decay method.

\begin{figure*}[t]
    \centering
    \includegraphics[width=\textwidth]{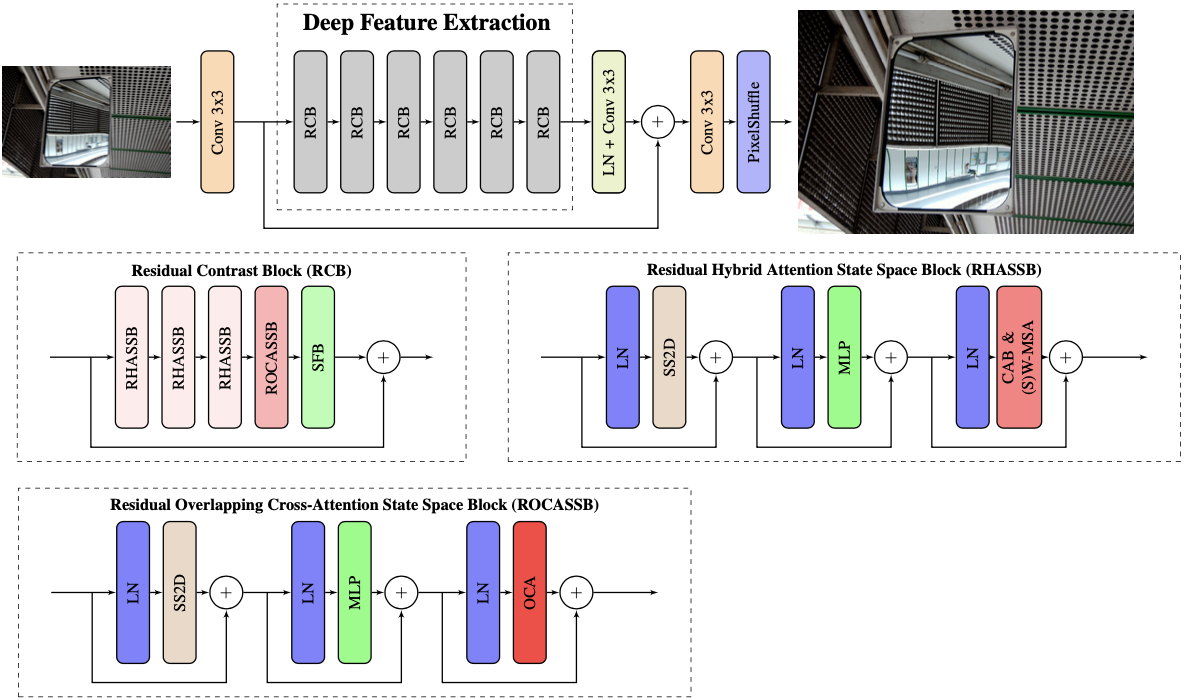}
    \vspace{-2.mm}
    \caption{\textbf{Team Contrast}}
    \label{fig:team13_01}
    \vspace{-4.mm}
\end{figure*}

\vspace{2.mm}
\subsection{BetterSR}
\paragraph{Description.}
The BetterSR team's method adopts the structure of the SwinFIR~\cite{zhang2022swinfir} for single image super-resolution. Building upon the foundations laid by SwinIR~\cite{liang2021swinir}, SwinFIR~\cite{zhang2022swinfir} introduces a novel Fast Fourier Convolution (FFC) operator. This allows SwinFIR~\cite{zhang2022swinfir} to capture global information more efficiently.
The overall architecture of the proposed solution is presented in \cref{fig:team12_01}. The main components are 5 Residual Swin Transformer Blocks (RSTB) and 1 Spatial Frequency Block (SFB)~\cite{zhang2022swinfir}. 
The proposed solution is trained through multiple stages, employing a greedy policy. The learning rate undergoes progressive decay by a factor of 0.1 after each stage. Within each training stage, the learning rate remains fixed, and the authors select the best-performing saved checkpoint as the pretrained model for subsequent stages.

\vspace{-4.mm}
\paragraph{Implementation Details.}
Building upon SwinFIR~\cite{zhang2022swinfir}, the proposed solution integrates a neural degradation algorithm~\cite{luo2024and} for data augmentation during training. Furthermore, the authors enrich the DIV2K training set~\cite{timofte2017ntire} with real-world paired data~\cite{guo2022data}, enhancing the diversity of the training set and improving the model's robustness against unseen data.
The authors utilize the official training code to retrain SwinFIR using their augmented training set. Besides, the training process is initiated by employing the official pretrained model `SwinFIR\_SRx4'.

\subsection{Contrast}
\paragraph{Description.} Drawing inspiration from the prominent State Space models like Mamba \cite{mamba,liu2024vmamba}, the prowess of the Hybrid Attention Transformer (HAT) \cite{chen2023activating}, and the application of Fast Fourier Transform (FFT) \cite{zhang2022swinfir}—each method celebrated for its exceptional performance—I embarked on a quest to synthesize these three formidable ideas into a neural network. This synthesis led to the inception of ``Contrast'', a neural network aptly named after its foundational pillars: Convolution, Transformer, and State Space. The Contrast model features a novel Residual Contrast Block (RCB), which is a conglomeration of distinct yet harmonious elements: the Residual Hybrid Attention State Space Block (RHASSB), the Residual Overlapping Cross-Attention State Space Block (ROCASSB), and the Spatial Fourier Block (SFB) inspired by the SwinFIR and SwinIR\cite{liang2021swinir} models. The RHASSB and ROCASSB share structural similarities but are distinguished by their unique attention blocks. The detailed architecture is delineated in \cref{fig:team13_01}.

\paragraph{Implementation Details.} The Contrast model's training strategy was grounded in a comprehensive dataset ensemble, incorporating DIV2K \cite{timofte2017ntire}, Flickr2K \cite{Lim_2017_CVPR_Workshops}, and LSDIR \cite{li2023lsdir}. To enrich the training set and enhance model robustness, various data augmentation techniques were applied, including random rotations, random flips, and color channel shuffling. The network was optimized using Mean Squared Error Loss (MSELoss) in conjunction with the Adam optimizer. An Exponential Moving Average (EMA) with a decay rate of 0.999 was implemented to stabilize the training updates. The learning rate was set at 2$\times$$10^{-4}$.

The model was trained on 64$\times$64 image patches, selected to balance detail capture and computational efficiency. The patchification size was 1, with a window size of 16 and an embedding dimension of 180, to facilitate a comprehensive feature extraction. Training was executed for 180,000 iterations with a batch size of 32. Due to time constraints, the network did not undergo extended training, yet even in its less-trained state, it demonstrated promising results, indicative of its potential in super-resolution tasks. The integration of the novel RCB, alongside these training parameters, underscores the model's scientific rigor and its application in advancing neural network design for image processing.

\begin{figure*}[th]
    \centering
    \includegraphics[width=\textwidth]{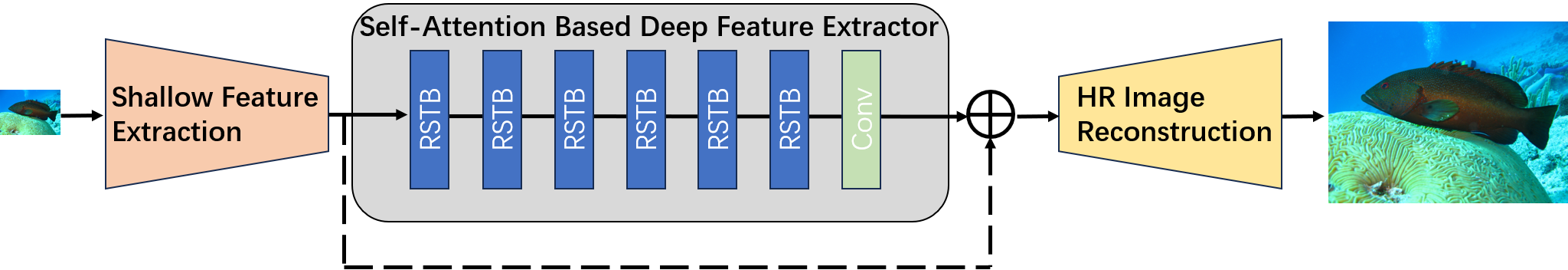}
    \caption{\textbf{Team BFU-SR}}
    \label{fig:team14_01}
\end{figure*}

\subsection{BFU-SR}
\paragraph{Description.}Drawing upon the concept of the Swin structure as mentioned in \cite{liu2021swin} and the idea of self-attention introduced in \cite{self_attention}, the authors employ SwinIR \cite{liang2021swinir} as their core network architecture, as shown in \cref{fig:team14_01}, distinguished by its modular design encompassing shallow feature extraction, deep feature extraction, and high-quality image reconstruction modules. The process begins with a low-resolution input image that undergoes initial processing through a convolutional layer in the shallow feature extraction module. This step is pivotal for mapping the input to a higher-dimensional feature space, facilitating early visual processing and promoting stable optimization. The architecture's uniqueness lies in its ability to leverage simple yet effective convolutional operations at this stage, setting a strong foundation for subsequent processing.

Subsequent deep feature extraction involves a series of residual Swin Transformer blocks followed by an additional convolutional layer. This innovative approach integrates the convolutional inductive bias into a Transformer-based model, significantly enhancing the network's ability to process spatial information. The sequential extraction of features through these blocks progressively refines the representation of the input image, capturing intricate details and textures.

In the realm of super-resolution (SR) tasks, the proposed strategy involves the fusion of shallow and deep features to reconstruct high-quality images. Shallow features, adept at capturing low-frequency information, ensure that the basic structure and smooth areas of the image are preserved. Meanwhile, deep features focus on reconstructing high-frequency details, critical for restoring sharp edges and fine textures. The reconstruction module employs a sub-pixel convolution layer for efficient upsampling, meticulously engineered to enhance the resolution while maintaining the integrity of the original image.

This holistic approach, leveraging both shallow and deep feature extraction, allows SwinIR to excel across a broad spectrum of image restoration tasks. By meticulously balancing the contribution of each feature type, the authors ensure that the reconstructed images achieve unparalleled clarity and detail, setting a new benchmark in the field of super-resolution and beyond. 

\textbf{Loss Function.} Building on the methodologies described in previous studies \cite{loss1,loss2,loss3}, the authors design and adopt a hybrid loss function as follows:
\begin{equation}
\mathcal{L}_{\text{total}} = \mathcal{L}_{\text{smooth}} + \alpha \mathcal{L}_{\text{MS-SSIM}} + \beta \mathcal{L}_{\text{per}},
\end{equation}
where $\alpha$$=$$0.1$ and $\beta$$=$$0.01$ are the hyperparameters weighting each loss component.

\begin{figure*}[t]
    \centering
    \includegraphics[width=\textwidth]{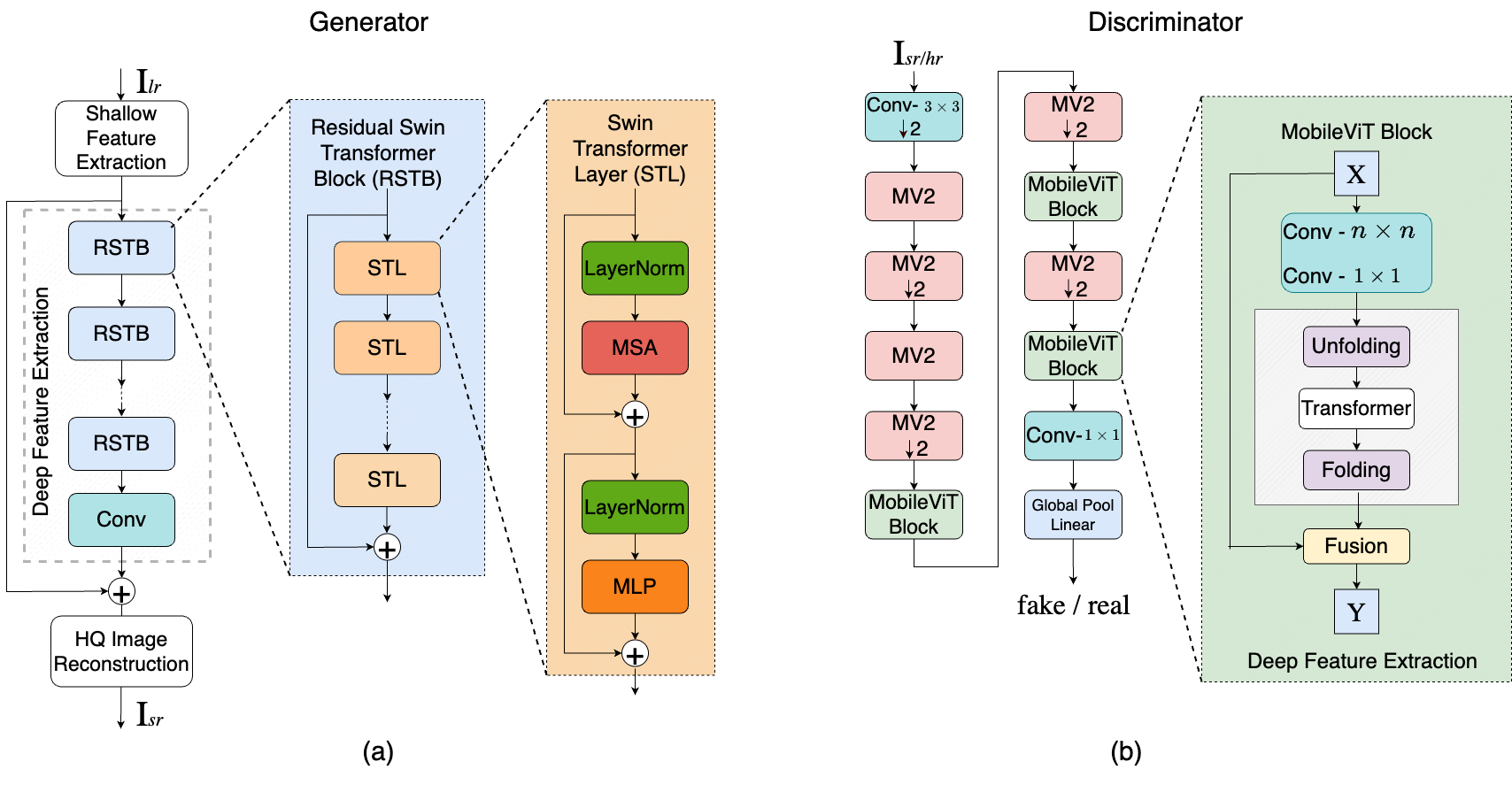}
    \vspace{-4.mm}
    \caption{\textbf{Team SCU-VIP-LAB}}
    \label{fig:team15_1}
\end{figure*}

\vspace{-4.mm}
\paragraph{Implementation Details.} For the SwinIR model, the authors set the hyperparameters as follows: window size to 8, embedding dimension to 180, number of heads to 6, and employ 'pixelshuffle' for the upsampler. Regarding the training strategy, the model undergoes training utilizing a combined dataset from DIV2K, Flickr2K, and LSDIR. The training regimen is bifurcated into two phases. In the initial phase, patches of size 64×64 are randomly cropped from the images. To enhance the diversity of the training data, the authors apply random rotations (90, 180, or 270 degrees) and horizontal flips. The settings for batch size, initial learning rate, and total iterations are configured to 32, $1$$\times$$10^{-4}$, and 1,000,000, respectively, with the learning rate being reduced by half after 600,000 iterations. In the subsequent phase, the model is fine-tuned using images of size 128$\times$128 over 100,000 iterations.

\begin{figure}[t]
    \centering
    \includegraphics[width=\columnwidth]{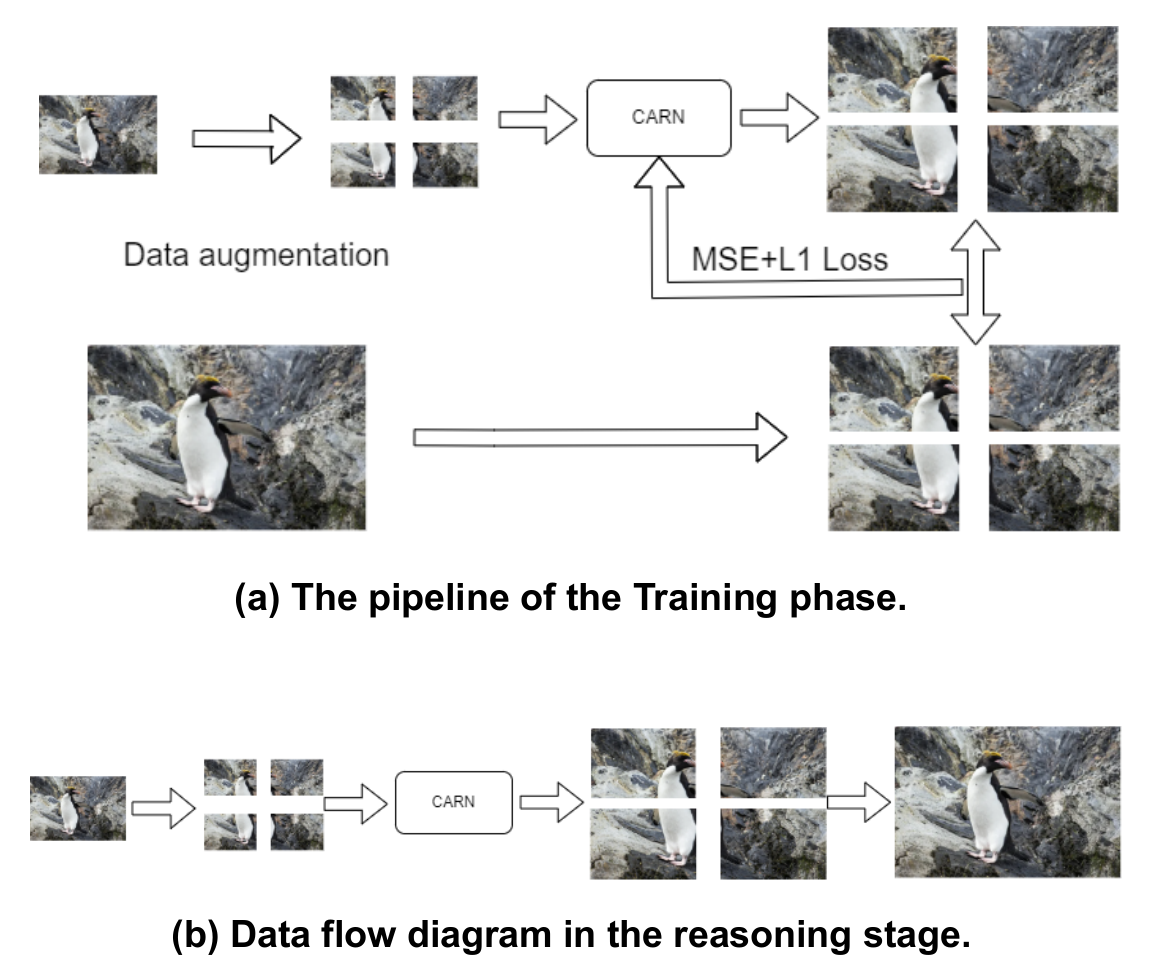}
    \vspace{-4.mm}
    \caption{\textbf{Team Nudter}}
    \vspace{-2.mm}
    \label{fig:team16_1}
\end{figure}

\begin{figure*}[t]
    \centering
    \includegraphics[width=\textwidth]{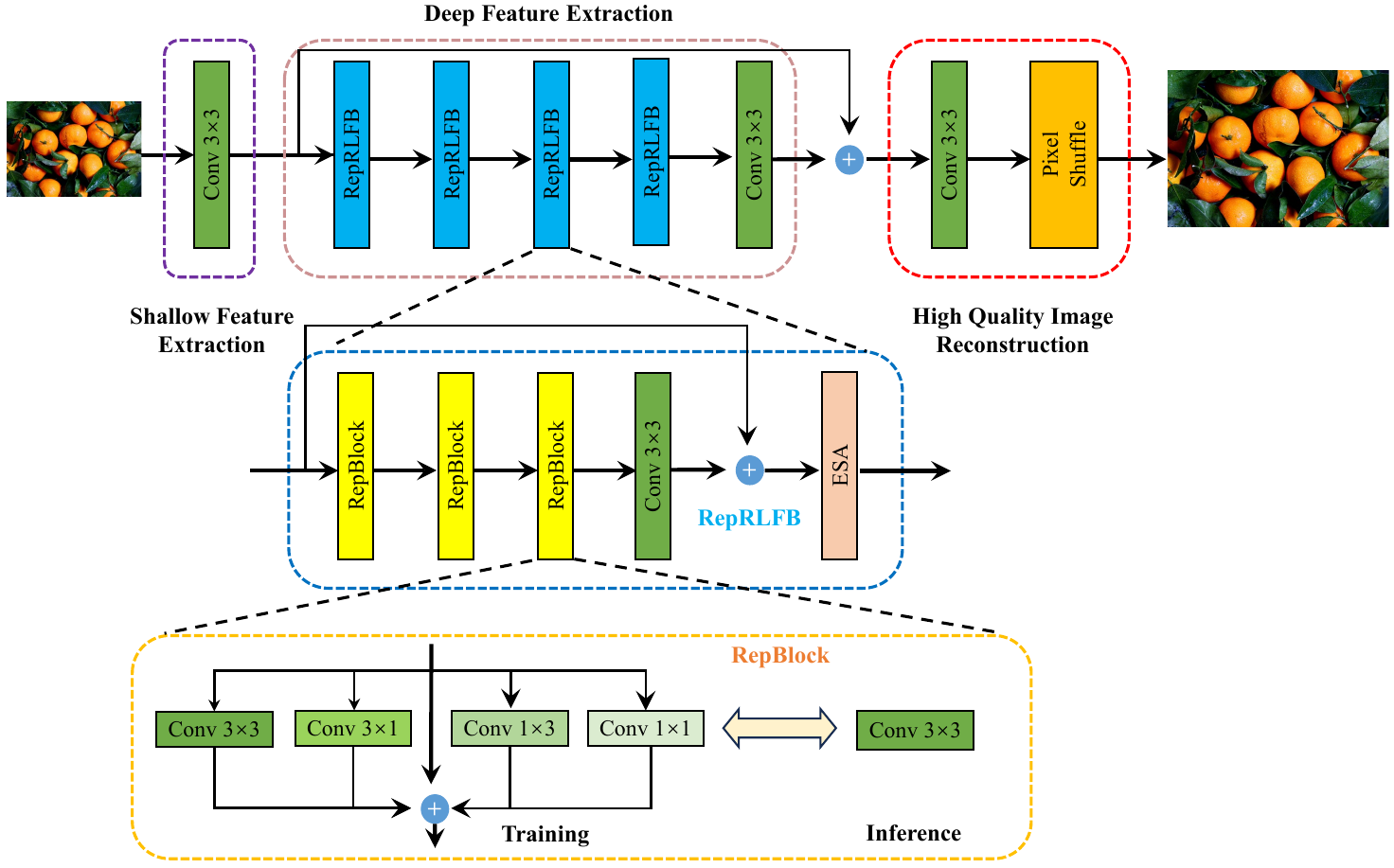}
    \caption{\textbf{Team JNU-620}}
    \label{fig:team17_01}
    \vspace{-2.mm}
\end{figure*}

\subsection{SCU-VIP-LAB}
\paragraph{Description.}
This team opted for utilizing a GAN for their model, inspired by the adversarial architecture in
\cite{sr360}. As shown in Fig. \ref{fig:team15_1} (a), the generator is the light-weight SwinIR
\cite{liang2021swinir}. It is made up of Residual Swin Transformer Blocks (RSTBs). RSTBs are
based on the original Swin Transformer Block architecture, which capture both local and global
information. The Swin Transformer Layer (STL) is a self-attention based architecture that breaks
down the image to patches, or shifted windows to process the information effectively. By
incorporating residual connections among the STL blocks, the RSTBs can mitigate the vanishing
gradient problem. This allows RSTBs to effectively train deeper architectures.

The authors' proposed discriminator is MobileViT \cite{mehta2021mobilevit}, a transformer-
based image classifier with a light-weight design, shown in \ref{fig:team15_1} (b). MobileViT
was developed from the Vision Transformer with resource-constrained devices in mind. Because
of this, it has fewer attention heads, smaller hidden embedding, and a more computationally
efficient attention mechanism. The light-weight SwinIR model uses the same methods to reduce
the model size starting from SwinIR used for classical image super-resolution. This proposed
GAN network, LiteSwinIRplus, attempts to balance efficiency and performance.

\vspace{-4.mm}
\paragraph{Implementation Details.}
During training, the authors initialized the generator using the pre-trained light-weight SwinIR
model from \cite{liang2021swinir}. The hyperparameters of SwinIR were tweaked to be more
efficient from a computation point of view and set as follows: window size to 8, embedding
dimension to 60, and the number of heads to 6. This generator was then trained with the
MobileViT discriminator for 71 epochs on 800 images from the DIV2K training set. The images
were trained in batches of 16 and randomly cropped to a size of $64$$\times$$64$. The generator
and discriminator were updated alternately using stochastic gradient descent. The learning rate
for updating the generator is 2$\times$$10^{-5}$, and the learning rate for updating the discriminator is 5$\times$$10^{-4}$.

\subsection{Nudter}
\paragraph{Description.}
The authors choose Cascading Residual Network (CARN) as the backbone of the model, leveraging its cascading residual structure to optimize feature transmission. To enhance the generalization ability of the model, they adopt a data enhancement strategy of dividing the image into four parts. At the same time, the model's loss function integrates MSE and L1 losses to comprehensively optimize performance.

The overall workflow of the solution proposed by their team is shown in~\cref{fig:team16_1}\textcolor{red}{a}. 

During the training phase, local data augmentation is performed on each image in the DIV2K training set to achieve a reduction in size and enhancement of details. In this process, each low-resolution image and its corresponding high-resolution version are evenly divided into four smaller blocks. This operation has two major advantages:

\begin{itemize}
\item This strategy not only multiplies the original training samples by four but also broadens the training data context, as each segment provides distinct image content and structural insights.
\item By downscaling image dimensions, the model is trained to refine details from local attributes, improving performance with small-scale data. Moreover, when the model encounters a complete high-resolution image, the ability to extract and utilize local features.
\end{itemize}

\vspace{2.mm}
During training, the mean squared error loss LMSE and L1 loss LL1 are calculated independently for each part, and the average of all parts is used to supervise the learning process of the model:
\begin{equation}
L_{\text {joint }}=L_{M S E}+\lambda L_{L 1},
\end{equation}
where the hyper-parameter $\lambda$ is used to balance the weights of two losses.

During the inference stage, the team first processes the original low-resolution image. To ensure both the efficiency of the processing and the integrity of the image details, the image is expertly cropped into four small blocks. Additionally, to prevent the loss of crucial edge information during the cropping process, the team determines an appropriate shave size, thus ensuring the integrity and continuity of the image edges.

Subsequently, these four small image blocks are individually and orderly input into the pre-trained CARN model to obtain the corresponding high-resolution images. Finally, these four high-resolution sub-images processed by the CARN model are stitched together to obtain the final high-resolution image.

Through this inference process, the original low-resolution image is successfully converted into a high-resolution image, significantly improving the clarity and visual effect of the image.The above process can be reflected in~\cref{fig:team16_1}\textcolor{red}{b}. 

\paragraph{Implementation Details.}
During the training phase, MSE loss and L1 loss are used for super-resolution reconstruction. The training of the aforementioned models is conducted on one RTX 3090 GPU, with the Adam optimizer and multi-step learning rate decay method being applied simultaneously. The initial learning rate is set to 1$\times$$10^{-4}$.

\subsection{JNU-620}
\paragraph{Description.}
Inspired by RLFN~\cite{kong2022residual} and RepRFN~\cite{deng2023reparameterized}, the authors introduce RepRLFN by employing structural re-parameterization technology~\cite{ding2019acnet,deng2023reparameterized} based on RLFN~\cite{kong2022residual}, as shown in \cref{fig:team17_01}. Known for its effectiveness in super-resolution tasks, HAT~\cite{chen2023activating} is utilized alongside RepRLFN for model ensemble, leading to promising results.

RepRLFN mirrors the architecture of RepRFN~\cite{deng2023reparameterized}, differing only in the substitution of RepRLFBs for RepRFBs. As a core element of RepRLFB, RepBlock employs parallel branch structures to extract features from diverse receptive fields and modes. Structural reparameterization technology is employed to mitigate computational complexity during the inference stage, as a solution to the potential increase in complexity due to multi-branch integration. 

HAT~\cite{chen2023activating} is structured into three segments: shallow feature extraction, deep feature extraction, and image reconstruction. By integrating channel attention and window-based self-attention mechanisms, HAT~\cite{chen2023activating} effectively activates pixels crucial for super-resolution reconstruction.

\paragraph{Implementation Details.}
To train RepRLFN, the diverse datasets comprising DIV2K, Flicker2K, and the initial 10k images of LSDIR were utilized. HR images were randomly cropped into $480 \times 480$ patches, with LR images cropped correspondingly. Data augmentation was applied including random horizontal/vertical flipping and RGB channel shuffling. Training began with L1 loss, followed by fine-tuning using MSE loss via Adam optimizer. 

For training HAT~\cite{chen2023activating}, DIV2K and Flickr2K datasets were employed. The L1 loss was utilized with a learning rate of $1$$\times$$10^{-5}$, alongside data augmentation involving horizontal flips and 90-degree rotations. A batch size of 2 and a patch size of 64 were employed. 

During testing, a test-time data ensemble approach was applied to enhance performance. Additionally, a weighted fusion of RepRLFN and HAT~\cite{chen2023activating} results produced the final output.

\begin{figure*}[ht]
    \centering
    \begin{subfigure}[b]{\columnwidth}
        \centering
        \includegraphics[width=\textwidth]{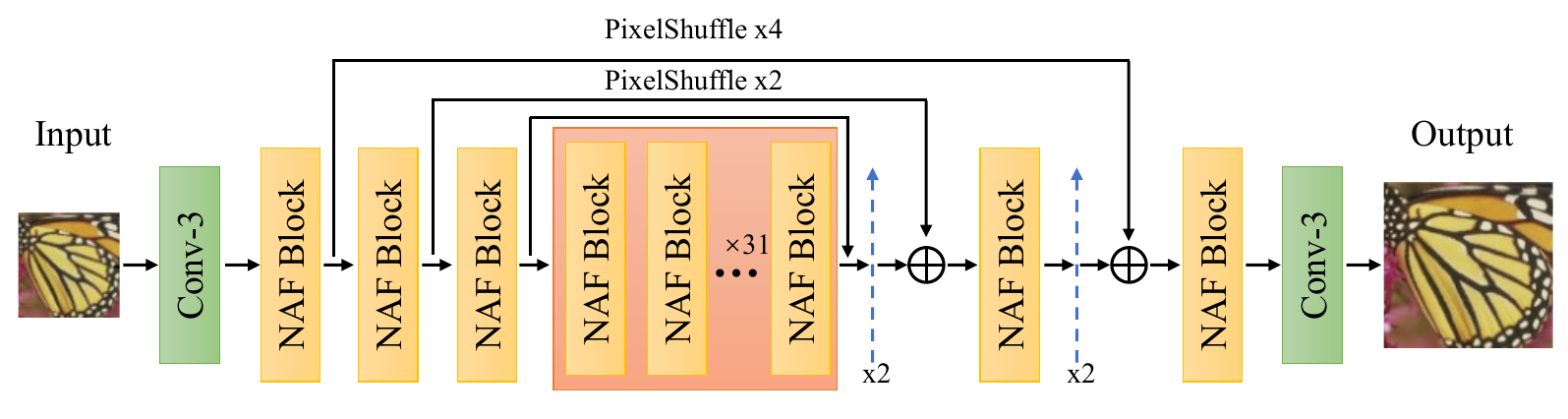}
        \subcaption{}
        \label{fig:team18_01}
    \end{subfigure}
    \hfill
    \begin{subfigure}[b]{\columnwidth}
        \centering
        \includegraphics[width=\textwidth]{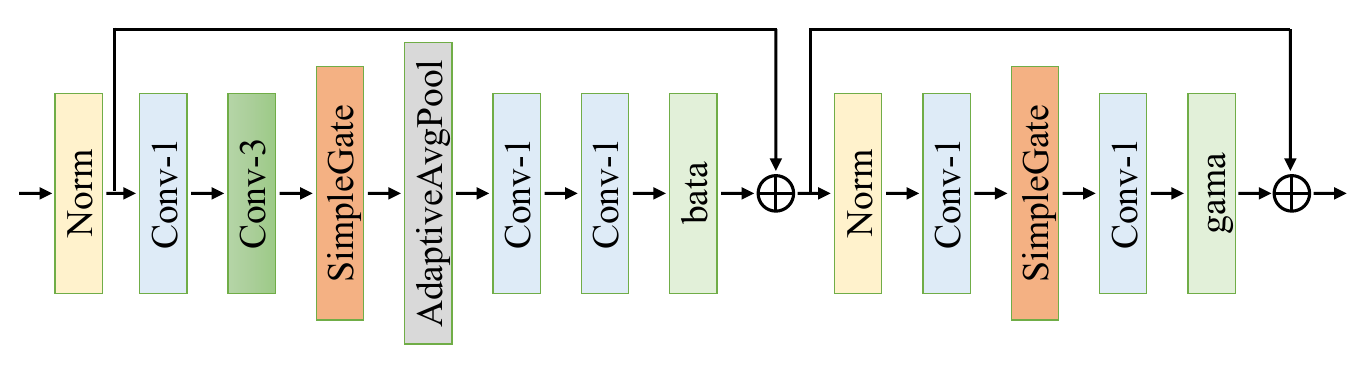}
        \subcaption{}
        \label{fig:team18_02}
    \end{subfigure}
    \vspace{-4.mm}
    \caption{\textbf{Team LVGroup-HFUT}}
    \label{fig:team18_subfigures}
    \vspace{2.mm}
\end{figure*}

\begin{figure*}[th]
    \centering
    \includegraphics[width=\textwidth]{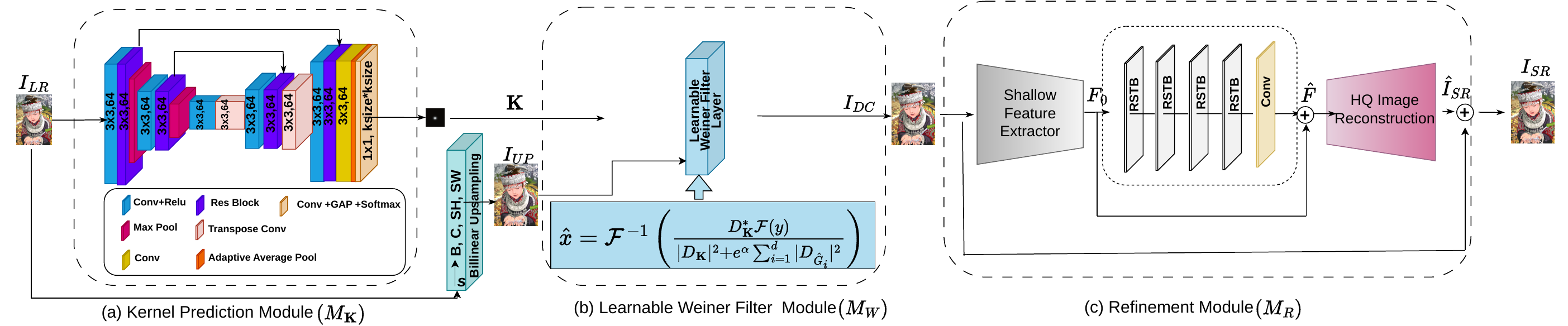}
    \vspace{-6.mm}
    \caption{\textbf{Team Uniud}}
    \label{fig:team19_01}
    \vspace{-2.mm}
\end{figure*}

\subsection{LVGroup-HFUT}
\paragraph{Description.}
Super-Resolution refers to the application of computational techniques to enhance the resolution of an image or video beyond its original captured quality. This field aims to reconstruct high-resolution (HR) images from their low-resolution (LR) counterparts. However, accurately inferring and reconstructing the fine details missing in the LR images based on the learned models of image textures and structures, poses a challenging task.

Therefore, the authors choose NAFNet~\cite{chen2022simple} with LPIPS perception constraint as their network architecture, as shown in Fig.~\ref{fig:team18_subfigures}. The input image will undergo processing through two phases: training and inference. During the training phase, the input image is fed into the NAFNet network, and it is constrained using three loss functions: lpips loss with weight 0.5, fft loss with weight 0.1, and L1 loss with weight 1.

The process begins with an initial convolutional layer (Conv-3) with a kernel size of $3$$\times$$3$. Following this, a series of Residual Feature Block (RFB) modules, labeled as 'RFB\_light', are employed. These blocks are typically designed to learn and refine feature representations. The model utilizes bilinear interpolation upsampling as a mechanism to increase the spatial resolution of the feature maps. Concurrently, the upscaled features are passed through another convolutional layer (Conv-3) to further refine the feature maps, ensuring they align with the increased resolution. These refined features are then fed into a PixelShuffle layer, a commonly used technique in super-resolution that reorganizes the feature map from the channel dimension into the spatial dimensions to achieve the final output.

\paragraph{Implementation Details.}
The proposed architecture is based on PyTorch 2.2.1 and an NVIDIA 4090 with 24G memory. the authors set 1500 epochs for training with batch size 32, using AdamW with $\beta_1$$=$$0.9$ and $\beta_2$$=$$0.999$ for optimization.
The initial learning rate was set to 0.001. For data augment, the authors first randomly crop the image to 48$\times$48 and then perform horizontal flip with probability 0.5. The image is preprocessed with a kernel number $64$ of the latent dim for scale up. During inference, the test images with original resolution are feed forward in the corresponding pretrained model. Training strategies: During the training phase, the input image is fed into the NAFNet network, and it is constrained using three loss functions: lpips loss with weight 0.5, fft loss with weight 0.1, and L1 loss with weight 1.

\begin{figure*}[ht]
    \centering
    \begin{subfigure}[b]{\columnwidth}
        \centering
        \includegraphics[width=\textwidth]{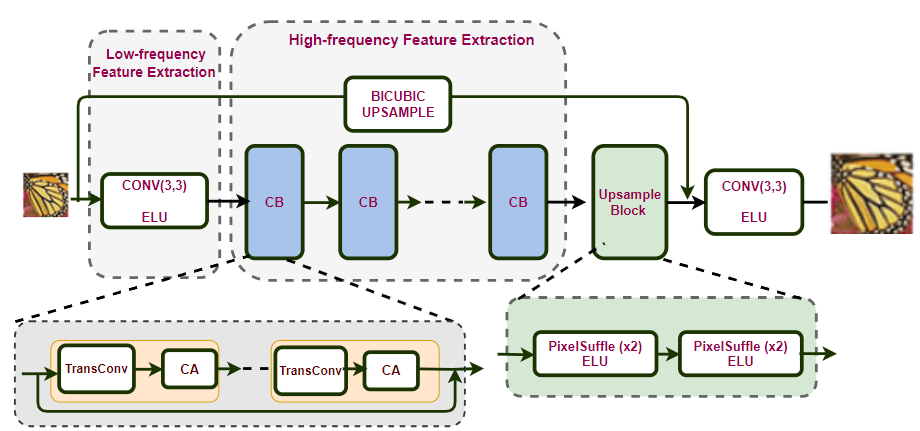}
        \subcaption{}
        \label{fig:team20_01}
    \end{subfigure}
    \hfill
    \begin{subfigure}[b]{\columnwidth}
        \centering
        \includegraphics[width=\textwidth]{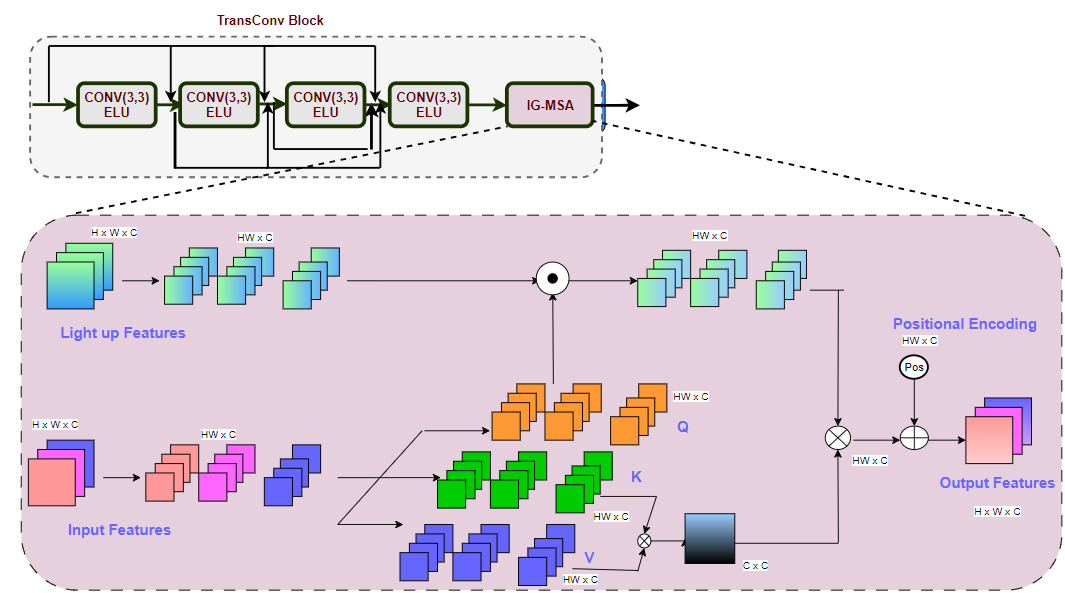}
        \subcaption{}
        \label{fig:team20_02}
    \end{subfigure}
    \caption{\textbf{Team SVNIT-NTNU-1}}
    \label{fig:team20_subfigures}
\end{figure*}

\vspace{4.mm}
\subsection{Uniud}
\paragraph{Description.} The authors propose a blind-SR approach, shown in \cref{fig:team19_01}, consisting of three main modules: the Kernel Prediction Module (KPM), the Learnable Wiener Filter Module (LWFM), and the Refinement Module (RM).
The KPM implicitly estimates the degradation kernel $\mathbf{K}$$\in$$\mathbb{R}^{k \times k}$ from the LR input $\mathbf{I}_{LR}$$\in$$\mathbb{R}^{C \times H\times W}$, where $C$ denotes the number of channels, $H$ and $W$ are spatial dimensions.
With a parallel stream, $\mathbf{I}_{LR}$ is then upsampled (via bilinear interpolation with scale factor $S$) to generate $\mathbf{I}_{UP}$$\in$$\mathbb{R}^{C \times SH \times SW }$.
$\mathbf{I}_{UP}$ and $\mathbf{K}$ are the inputs to the LWFM.
the authors derived a novel formulation within such a module that leverages efficient operation in the frequency domain to learn the Wiener filter parameters with a closed-form solution.
This computationally efficient approach yields to $\mathbf{I}_{DC}$$\in$$\mathbb{R}^{C \times SH \times SW}$ that the RM finally exploits to generate the SR image $\mathbf{I}_{SR}$$\in$$\mathbb{R}^{C \times SH \times SW}$.

\vspace{4.mm}
\textbf{Optimization}
The model is trained using an end-to-end strategy optimizing
\begin{equation}
\label{eq:total_loss}
\mathcal{L}_{total} = \lambda_{1}\mathcal{L}_{SR} + \lambda_{2}\mathcal{L}_{k} + \lambda_{3}\mathcal{L}_{TV},
\end{equation}
where
\begin{equation}
\label{eq:l1_loss}
\mathcal{L}_{SR} = \|\mathbf{I}_{HR} - \mathbf{I}_{SR} \|_{1},
\end{equation}
is the image reconstruction loss quantifying the accuracy of the super-resolved output $\mathbf{I}_{SR}$ against the HR ground-truth $\mathbf{I}_{HR}\in \mathbb{R}^{C \times SH \times SW}$.

The novel kernel estimation loss, $\mathcal{L}_{k}$, is a key component in their work.
Considering the blurring process assumption, the authors designed the function
\begin{equation}
\label{eq:ski_loss}
\mathcal{L}_{k} = 
\|\mathbf{I}_{LR} - \left( \mathbf{I}_{HR} \circledast M_{\mathbf{\hat{K}}}(\mathbf{I}_{LR}) \right)\downarrow_{S}\|_{1},
\end{equation}
which, through the L1 penalty, forces the model to learn a kernel that convolved with the ground truth HR would generate the same LR input sample, thus bypassing the need for explicit kernel definition.

To complement this, the total variation loss
\begin{equation}
\label{eq:tv_loss}
\mathcal{L}_{TV} = \|\triangledown \mathbf{I}_{HR} - \triangledown \mathbf{I}_{SR} \|_{1},
\end{equation}
computes the difference between horizontal and vertical gradients (denoted with $\triangledown$) to encourage smoothness of the image by minimizing the variations in pixel intensities, hence
reducing noise and unwanted artifacts.
$\lambda_{1}$, $\lambda_{2}$ and $\lambda_{3}$ are balancing parameters.

\paragraph{Implementation Details.}
The authors trained their model with $32$$\times$$32$ LR image patches for 500k iterations.
the authors used a batch size of 16 with the Adam optimizer~\cite{kingma2014adam} having $\beta_{1}$$=$$0.9$, $\beta_{2}$$=$$0.999$, and~$\epsilon$$=$$10^{-8}$. 
the authors set the learning rate to $10^{-4}$, then reduced it by a factor of 2 after 250k, 400k, 450k, and 475k iterations.
All $\mathrm{Conv2D}$ layers and residual blocks in the model have $3 \times 3$ kernels producing 64 output feature maps, except for HQ image reconstruction emitting 3 feature maps.
Within the RM module, the authors use $\gamma=4$ RSTB blocks, each composed of $6$ STL layers~\cite{liang2021swinir} with $96$ feature maps.
The LWFM considered $d$$=$$24$ with a learnable wiener filter size of $5$$\times$$5$~\cite{umer2019deep}, whose weights are initialized using the discrete cosine transform (DCT) basis.
Random vertical and horizontal flipping and $90°$~rotations were used as data augmentation strategies.The experiments were executed on an Intel Xeon processor, with 188 GB of RAM, and an NVIDIA A100 on Ubuntu 20.04 LTS implemented in Pytorch.

\subsection{SVNIT-NTNU-1}
\paragraph{Description.}  In order to design single image super-resolution, the authors use dense convolutional neural network approach in the proposed solution.
Model architecture description. The \cref{fig:team20_01} depicts the proposed architecture for single image super-resolution for scaling factors of $\times4$. The LR image is applied as input to the network and it is passed to extract the salient features from it. The low frequency features are extracted with first layers that employ convolutional layer While high frequency features are extracted with Concatenate Blocks (CB). The architecture uses the Exponential Linear Unit (ELU) activation function to improve learning performance at each layer in efficient manner. A new and core element of the proposed architecture is the partially densely connected design of TransConv blocks that preserves the high frequency details of the SR image by retaining salient features which is displayed \cref{fig:team20_02}. The kernel sizes (\ie, $3\times3$ ) adopted in TransConv blocks recover details distributed at local and global regions. Light-up feature is fed to IG-MSA which treats a single-channel feature map as a token and computes the self attention \cite{igmsa}. The channel attention modules are further used in Concanate block to perform adaptive re-scaling of features on per-channel basis. The pixel Shuffle is used to upscale the feature maps to the desired scaling factor (\ie, $\times4$) \cite{yang2019deep}.
Additionally, low frequency features are upscaled and added with the reconstructed output to retain more versatile information.

\paragraph{Implementation Details.}
The code is implemented using Pytorch library. The proposed network is trained using weighted combination of $l_1$, SSIM loss, Total Variation loss and charbornnier loss with a learning rate of $1$$\times$$10^{-4}$ which is decayed by $1$$\times$$10^4$ iterations and the same is optimized using Adam optimizer. The model is trained up to $1$$\times$$10^5$ iterations with a batch size of $8$.

\appendix
\section{Teams and Affiliations}
\label{sec:app_teams}
\subsection*{NTIRE 2024 team}
\noindent\textit{\textbf{Title: }} NTIRE 2024 Image Super-Resolution ($\times$4) Challenge\\
\noindent\textit{\textbf{Members: }} \\
Zheng Chen$^1$ (\href{mailto:zhengchen.cse@gmail.com}{zhengchen.cse@gmail.com}),\\
Zongwei Wu$^2$ (\href{zongwei.wu@uni-wuerzburg.de}{zongwei.wu@uni-wuerzburg.de}),\\
Eduard-Sebastian Zamfir$^2$ (\href{Eduard-Sebastian.Zamfir@uni-wuerzburg.de}{Eduard-Sebastian.Zamfir@uni-wuerzburg.de}),\\
Kai Zhang$^3$ (\href{mailto:cskaizhang@gmail.com}{cskaizhang@gmail.com}),\\
Yulun Zhang$^1$ (\href{mailto:yulun100@gmail.com}{yulun100@gmail.com}),\\
Radu Timofte$^2$ (\href{mailto:radu.timofte@uni-wuerzburg.de}{radu.timofte@uni-wuerzburg.de}),\\
Xiaokang Yang$^1$ (\href{mailto:xkyang@sjtu.edu.cn}{xkyang@sjtu.edu.cn})\\
\noindent\textit{\textbf{Affiliations: }}\\
$^1$ Shanghai Jiao Tong University, China\\
$^2$ University of W\"urzburg, Germany\\
$^3$ Computer Vision Lab, ETH Zurich, Switzerland\\

\subsection*{XiaomiMM}
\noindent\textit{\textbf{Title: }} MambaSR: State Space Transformer Model for Image Super Resolution\\
\noindent\textit{\textbf{Members: }} \\
Hongyuan Yu$^1$ (\href{mailto:yuhyuan1995@gmail.com}{yuhyuan1995@gmail.com}),
Cheng Wan$^2$,
Yuxin Hong$^3$,
Zhijuan Huang$^1$,
Yajun Zou$^1$,
Yuan Huang$^1$,
Jiamin Lin$^1$,
Bingnan Han$^1$,
Xianyu Guan$^1$,
Yongsheng Yu$^4$,
Daoan Zhang$^4$,
Xuanwu Yin$^1$,
Kunlong Zuo$^1$

\noindent\textit{\textbf{Affiliations: }}\\
$^1$ Multimedia Department, Xiaomi Inc\\
$^2$ Georgia Institute of Technology\\
$^3$ Lanzhou University\\
$^4$ University of Rochester\\

\subsection*{SUPSR}
\noindent\textit{\textbf{Title: }} SUPSR: Scaling Up Dataset for Image Super-Resolution\\
\noindent\textit{\textbf{Members: }} \\
Jinhua Hao$^1$ (\href{haojinhua206@gmail.com}{haojinhua206@gmail.com}),
Kai Zhao$^1$,
Kun Yuan$^1$,
Ming Sun$^1$, 
Chao Zhou$^1$ 

\noindent\textit{\textbf{Affiliations: }}\\
$^1$ Kuaishou Technology\\

\subsection*{UCAS-SCST}
\noindent\textit{\textbf{Title: }} HFT: High Frequency Transformer for Image Super-Resolution\\
\noindent\textit{\textbf{Members: }} \\
Hongyu An$^1$ (\href{anhongyu22@mails.ucas.ac.cn}{anhongyu22@mails.ucas.ac.cn}),
Xinfeng Zhang$^1$

\noindent\textit{\textbf{Affiliations: }}\\
$^1$ School of Computer Science and Technology, University of Chinese Academy of Sciences, China\\

\subsection*{SYSU-SR}
\noindent\textit{\textbf{Title: }} Pre-trained Model with Ensemble Learning for Image Super-resolution\\
\noindent\textit{\textbf{Members: }} \\
Zhiyuan Song$^1$ (\href{songzhy29@mail2.sysu.edu.cn}{songzhy29@mail2.sysu.edu.cn}),
Ziyue Dong$^2$,
Qing Zhao$^1$,
Xiaogang Xu$^3$,
Pengxu Wei$^1$

\noindent\textit{\textbf{Affiliations: }}\\
$^1$ Sun Yat-sen University, China\\
$^2$ Xi'an Jiaotong University, China\\
$^3$ Zhejiang University, China\\

\subsection*{Jasmine}
\noindent\textit{\textbf{Title: }} Hybrid Multi-Axis Aggregation Network for Image Super-Resolution\\
\noindent\textit{\textbf{Members: }} \\
Zhi-chao Dou$^1$ (\href{douzhichao2021@163.com}{douzhichao2021@163.com}),
Gui-ling Wang$^1$

\noindent\textit{\textbf{Affiliations: }}\\
$^1$ Shandong University of Science and Technology Qingdao, China\\

\subsection*{ACVLAB}
\noindent\textit{\textbf{Title: }} Self-ensemble Fusion Solution for Image Super-resolution\\
\noindent\textit{\textbf{Members: }} \\
Chih-Chung Hsu$^1$ ({\href{mailto:cchsu@gs.ncku.edu.tw}{cchsu@gs.ncku.edu.tw}),
Chia-Ming Lee$^1$,
Yi-Shiuan Chou$^1$

\noindent\textit{\textbf{Affiliations: }}\\
$^1$ Institute of Data Science, National Cheng Kung University, Taiwan\\

\subsection*{mandalinadagi}
\noindent\textit{\textbf{Title: }} Wavelettention: Wavelet-Domain Losses Elevate Hybrid Transformer Model for Image Super-Resolution\\
\noindent\textit{\textbf{Members: }} \\
Cansu Korkmaz$^1$ (\href{mailto:ckorkmaz14@ku.edu.tr}{ckorkmaz14@ku.edu.tr}),
A. Murat Tekalp$^1$

\noindent\textit{\textbf{Affiliations: }}\\
$^1$ Koc University\\

\subsection*{SKDADDYS}
\noindent\textit{\textbf{Title: }} EHAT: Enhanced Attention Transformer for Image Restoration\\
\noindent\textit{\textbf{Members: }} \\
Yubin Wei$^1$ (\href{mailto:yubinwei@stu.xmu.edu.cn}{yubinwei@stu.xmu.edu.cn}),
Xiaole Yan$^1$,
Binren Li$^1$,
Haonan Chen$^1$,
Siqi Zhang$^1$,
Sihan Chen$^1$

\noindent\textit{\textbf{Affiliations: }}\\
$^1$ Xiamen University\\

\subsection*{KLETech-CEVI-Lowlight-Hypnotise}
\noindent\textit{\textbf{Title: }{SR4X: Towards Image Super-Resolution ($\times$4)}}\\ 
\noindent\textit{\textbf{Members: }} \\ 
Amogh Joshi$^1$ (\href{mailto:joshiamoghmukund@gmail.com}{joshiamoghmukund@gmail.com}),
Nikhil Akalwadi$^{1,3}$,
Sampada Malagi$^{1,3}$,
Palani Yashaswini$^{1,2}$,
Chaitra Desai$^{1,3}$,
Ramesh Ashok Tabib$^{1,2}$,
Ujwala Patil$^{1,2}$,
Uma Mudenagudi$^{1,2}$

\noindent\textit{\textbf{Affiliations: }}\\
$^1$ Center of Excellence in Visual Intelligence (CEVI), KLE Technological University, Hubballi, Karnataka, INDIA\\ 
$^2$ School of Electronics and Communication Engineering, KLE Technological University, Hubballi, Karnataka, INDIA\\ 
$^3$ School of Computer Science and Engineering, KLE Technological University, Hubballi, Karnataka, INDIA\\

\subsection*{SVNIT-NTNU}
\noindent\textit{\textbf{Title: }} Hybrid Attention based Single Image Super-Resolution\\
\noindent\textit{\textbf{Members: }} \\
Anjali Sarvaiya$^1$ (\href{mailto:anjali.sarvaiya.as@gmail.com}{anjali.sarvaiya.as@gmail.com}),
Pooja Choksy1$^1$, 
Jagrit Joshi1$^1$,
Shubh Kawa1$^1$,
Kishor Upla1$^1$,
Sushrut Patwardhan2$^2$,
Raghavendra Ramachandra2$^2$

\noindent\textit{\textbf{Affiliations: }}\\
$^1$ Sardar Vallabhbhai National Institute of Technology, India\\
$^2$ Norwegian University of Science and Technology, Norway\\

\subsection*{ResoRevolution}
\noindent\textit{\textbf{Title: }} Enhanced Hybrid Attention Transformer\\
\noindent\textit{\textbf{Members: }} \\
Sadat Hossain$^1$ (\href{mailto:sadat@deltax.ai}{sadat@deltax.ai}),
Geongi Park$^1$,
S. M. Nadim Uddin$^1$

\noindent\textit{\textbf{Affiliations: }}\\
$^1$ DeltaX, Seoul, South Korea\\

\subsection*{BetterSR}
\noindent\textit{\textbf{Title: }} Augmented SwinFIR\\
\noindent\textit{\textbf{Members: }} \\
Hao Xu$^1$ (\href{mailto:xu338@mcmaster.ca}{xu338@mcmaster.ca}),
Yanhui Guo$^1$,

\noindent\textit{\textbf{Affiliations: }}\\
$^1$ McMaster University Hamilton, ON, Canada\\

\subsection*{Contrast}
\noindent\textit{\textbf{Title: }} Contrast: Marrying Convolutions, Transformers and State Space for Image Super-Resolution\\
\noindent\textit{\textbf{Members: }} \\
Aman Urumbekov$^1$ (\href{mailto:amanurumbekov@gmail.com}{amanurumbekov@gmail.com}),

\noindent\textit{\textbf{Affiliations: }}\\
$^1$ Kyrgyz State Technical University(KSTU), Kyrgyzstan\\

\subsection*{BFU-SR}
\noindent\textit{\textbf{Title: }} Single Image Super-resolution Using Swin Transformer\\
\noindent\textit{\textbf{Members: }} \\
Xingzhuo Yan$^1$ (\href{mailto:xingzhuo.yan@cn.bosch.com}{xingzhuo.yan@cn.bosch.com}),
Wei Hao$^1$,
Minghan Fu$^1$

\noindent\textit{\textbf{Affiliations: }}\\
$^1$ Bosch Investment Ltd.\\
$^2$ Fortinet, Inc.\\
$^3$ University of Saskatchewan\\

\subsection*{SCU-VIP-LAB}
\noindent\textit{\textbf{Title: }} Generative Adversarial Network LiteSwinIRplus Using SwinIR and MobileViT\\
\noindent\textit{\textbf{Members: }} \\
Isaac Orais$^1$ (\href{mailto:iorais@scu.edu}{iorais@scu.edu}),
Samuel Smith$^1$,
Ying Liu$^1$

\noindent\textit{\textbf{Affiliations: }}\\
$^1$ Santa Clara University, Santa Clara, California, USA\\

\subsection*{Nudter}
\noindent\textit{\textbf{Title: }} CARN-Based Joint Asymmetric Loss and Data Augmentation for Image Super-Resolution\\
\noindent\textit{\textbf{Members: }} \\
Wangwang Jia$^1$ (\href{mailto:1903388692@qq.com}{1903388692@qq.com}),
Qisheng Xu$^1$,
Kele Xu$^1$

\noindent\textit{\textbf{Affiliations: }}\\
$^1$ National University of Defense Technology, China\\

\subsection*{JNU-620}
\noindent\textit{\textbf{Title: }} Image Super-Resolution Reconstruction Using RepRLFN and HAT\\
\noindent\textit{\textbf{Members: }} \\
Weijun Yuan$^1$ (\href{mailto:yweijun@stu2022.jnu.edu.cn}{yweijun@stu2022.jnu.edu.cn}),
Zhan Li$^1$,
Wenqin Kuang$^1$,
Ruijin Guan$^1$,
RutingDeng$^1$

\noindent\textit{\textbf{Affiliations: }}\\
$^1$ Jinan University, China\\

\subsection*{LVGroup-HFUT}
\noindent\textit{\textbf{Title: }} Light-weight Nonlinear Activation Free Network for Image Super-Resolution\\
\noindent\textit{\textbf{Members: }} \\
Zhao Zhang$^1$ (\href{mailto:cszzhang@gmail.com}{cszzhang@gmail.com}),
Bo Wang$^1$,
Suiyi Zhao$^1$,
Yan Luo$^1$,
Yanyan Wei$^1$

\noindent\textit{\textbf{Affiliations: }}\\
$^1$ Hefei University of Technology, China\\

\subsection*{Uniud}
\noindent\textit{\textbf{Title: }} IDENet: Implicit Degradation Estimation Network for Efficient Blind Super Resolution\\
\noindent\textit{\textbf{Members: }} \\
Asif Hussain Khan$^1$ (\href{mailto:khan.asifhussain@spes.uniud.it}{khan.asifhussain@spes.uniud.it}),
Christian Micheloni$^1$,
Niki Martinel$^1$

\noindent\textit{\textbf{Affiliations: }}\\
$^1$ University of Udine, Italy\\

\subsection*{SVNIT-NTNU-1}
\noindent\textit{\textbf{Title: }} Trans-Conv based Single Image Super-Resolution\\
\noindent\textit{\textbf{Members: }} \\
Pooja Choksy$^1$ (\href{mailto:ds22ec003@eced.svnit.ac.in}{ds22ec003@eced.svnit.ac.in}),
Anjali Sarvaiya1$^1$,
Jagrit Joshi1$^1$,
Shubh Kawa1$^1$,
Kishor Upla1$^1$,
Sushrut Patwardhan$^2$,
Raghavendra Ramachandra$^2$

\noindent\textit{\textbf{Affiliations: }}\\
$^1$ Sardar Vallabhbhai National Institute of Technology, India\\
$^2$ Norwegian University of Science and Technology, Norway\\

{\small
\bibliographystyle{ieeenat_fullname}
\bibliography{egbib}
}

\end{document}